\documentclass[10pt,twocolumn,letterpaper]{article}

\usepackage[pagenumbers]{cvpr}
\newcommand{\pub}[1]{\textsuperscript{#1}}%
\definecolor{cvprblue}{rgb}{0.21,0.49,0.74}
\usepackage[pagebackref,breaklinks,colorlinks,allcolors=cvprblue]{hyperref}
\usepackage{bm}
\usepackage{amsmath}
\usepackage{algorithm}
\usepackage{algorithmic}
\usepackage{multicol,multirow}
\usepackage{hyperref}
\usepackage{tocloft}
\usepackage{titletoc}
\usepackage[table]{xcolor}

\def\paperID{5883} 
\def\confName{CVPR}
\def\confYear{2025}

\begin{document}
\title{\textit{CoMBO}: Conflict Mitigation via Branched Optimization for\\ Class Incremental Segmentation}

\author{
Kai Fang$^{1}$\thanks{Equal contribution.}\hspace{0.1cm}, 
Anqi Zhang$^{1}$\footnotemark[1]\hspace{0.1cm},
Guangyu Gao$^{1}$\thanks{Corresponding author, guangyugao@bit.edu.cn.
}\hspace{0.1cm},
Jianbo Jiao$^{2}$,
Chi Harold Liu$^{1}$,
Yunchao Wei$^{3}$\\
$^{1}$Beijing Institute of Technology\qquad 
$^{2}$University of Birmingham\qquad
$^{3}$Beijing Jiaotong University
}

\maketitle

\begin{abstract}
Effective Class Incremental Segmentation (CIS) requires simultaneously mitigating catastrophic forgetting and ensuring sufficient plasticity to integrate new classes. 
The inherent conflict above often leads to a back-and-forth, which turns the objective into finding the balance between the performance of previous~(old) and incremental~(new) classes.
To address this conflict, we introduce a novel approach, Conflict Mitigation via Branched Optimization~(CoMBO).
Within this approach, we present the Query Conflict Reduction module, designed to explicitly refine queries for new classes through lightweight, class-specific adapters.
This module provides an additional branch for the acquisition of new classes while preserving the original queries for distillation. 
Moreover, we develop two strategies to further mitigate the conflict following the branched structure, 
\textit{i.e.}, the Half-Learning Half-Distillation~(HDHL) over classification probabilities, and the Importance-Based Knowledge Distillation~(IKD) over query features.
HDHL selectively engages in learning for classification probabilities of queries that match the ground truth of new classes, while aligning unmatched ones to the corresponding old probabilities, thus ensuring retention of old knowledge while absorbing new classes via learning negative samples.
Meanwhile, IKD assesses the importance of queries based on their matching degree to old classes, prioritizing the distillation of important features and allowing less critical features to evolve.
Extensive experiments in Class Incremental Panoptic and Semantic Segmentation settings have demonstrated the superior performance of CoMBO. 
Project page: \url{https://guangyu-ryan.github.io/CoMBO}.
\end{abstract}

\section{Introduction}
\label{sec:intro}
Semantic segmentation, the fundamental task in computer vision, involves classifying each pixel into predefined categories.
Panoptic segmentation, the challenging variant, unifies semantic and instance segmentation to both classify every pixel and simultaneously identify distinct object instances.
Recent advancements in mask classification-based segmentation, such as MaskFormer~\cite{maskformer} and Mask2Former~\cite{mask2former}, enable the unification of semantic and panoptic segmentation.
\begin{figure}
    \centering
    \includegraphics[width=\linewidth]{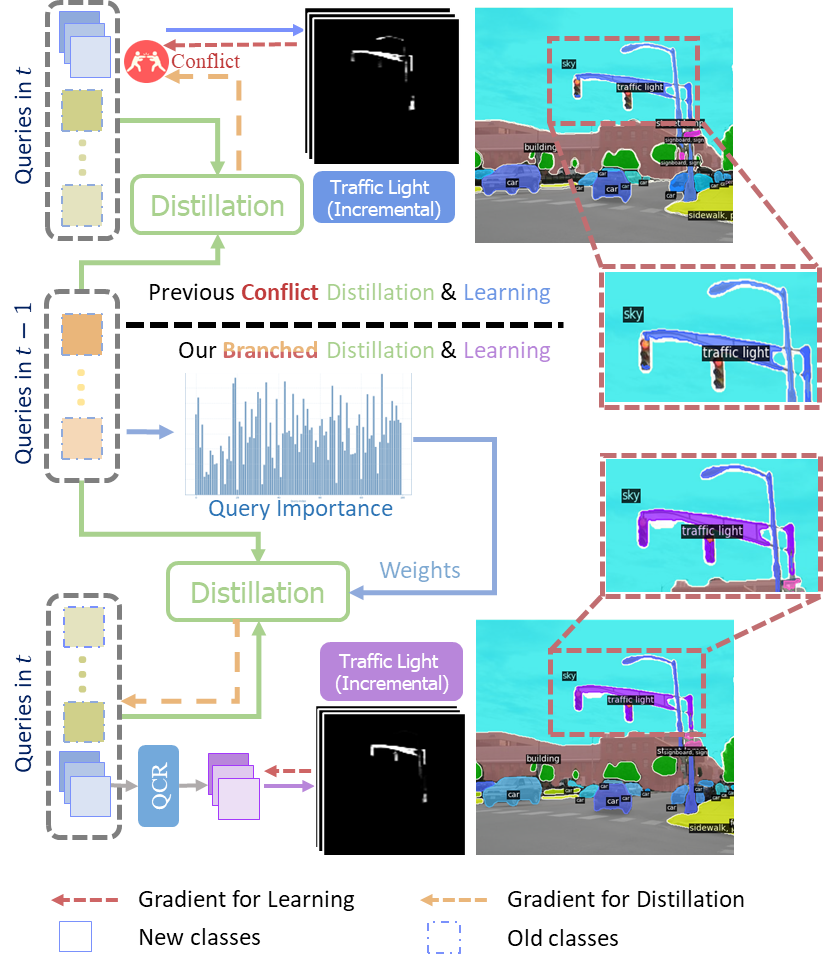}
    \caption{Comparison of our improved distillation and learning strategy (bottom) with previous conflicting strategy (top). 
    Previous strategies impose contradictory supervision on the same target to find a balance, whereas our strategy, including the importance factor for distillation and the QCR module for adaptive learning, enables more compatible, target-specific supervision. 
    The visualization on incremental classes (`\textit{Traffic Light}') highlights the effectiveness of our strategy.}
    \label{fig:brief}
    \vspace{-0.3cm}
\end{figure}
However, traditional approaches to segmentation suffer from limitations in dynamic environments where new classes can emerge unpredictably, as they are typically trained on a static set of categories.
Therefore, Class Incremental Semantic Segmentation (CISS) has emerged as a pivotal task for continually learning new categories, where models can adapt to new classes without forgetting previously learned ones.

Historically, studies in CISS~\cite{alife22,dkd22,coinseg23} mainly focuses on tackling \textit{catastrophic forgetting}, employing various techniques such as knowledge distillation~\cite{mib20,plop21,rcil22}, freezing parameters~\cite{ssul21,micro22,eclipse24}, or generating pseudo-labels from old model~\cite{plop21}. 
More recent efforts~\cite{comformer23,balconpas24} 
have extended these techniques to the more challenging Class Incremental Panoptic Segmentation~(CIPS) based on the advanced Mask2Former.
Nonetheless, these methods often fail to adequately reconcile the inherent tension between retaining old knowledge and acquiring new information.
Some overemphasize maintaining stability, thus hindering the learning of new classes, while others prioritize plasticity, leading to significant forgetting of old classes. 
It is necessary to deconstruct these conflicting objectives to facilitate a more compatible incremental learning process, achieving promotion in both the acquisition of new classes and retention of old classes.

In this work, we propose Conflict Mitigation via Branched Optimization~(CoMBO), an approach to reduce conflicts between acquiring new classes and retaining old classes.
By integrating three specialized knowledge distillation mechanisms within the Mask2Former architecture, CoMBO effectively mitigates catastrophic forgetting while enhancing the assimilation of new classes.
Using bipartite matching, Mask2Former optimally aligns each query, complete with its classification logits and mask predictions, with the most fitting ground truth class, ensuring efficient optimization.
Therefore, we first introduce the Half-Learning Half-Distillation mechanism, which selectively applies a shared Kullback-Leibler divergence to the classification logits of unmatched queries.
This method distinctly separates distillation on unmatched masks from the classification optimization of matched ones.
Moreover, rather than applying uniform distillation across all features, we introduce an importance-based knowledge distillation technique. 
This technique assesses each query's significance in preserving knowledge of old classes and adjusts the distillation intensity accordingly, emphasizing the retention of key queries.
Additionally, our Query Conflict Reduction module innovatively adapts features for new classes while maintaining the queries' original characteristics before distillation, using class-specific adapters for each newly learned class.
Finally, our approach outperforms previous methods in multiple benchmarks, especially in typical and challenging scenarios such as 100-10 of ADE20K. 
Our approach achieves remarkable performance of $35.6\%$ and $41.1\%$ mIoU on 100-10 of CIPS and CISS, respectively, compared to the previous state-of-the-art.

Overall, our contributions are summarized as follows:
\begin{itemize}
    \item We propose a Half-Learning Half-Distillation mechanism that applies soft distillation for old classes and binary-based optimization for new classes to effectively balance retention and acquisition.
    \item An extra Query Conflict Reduction module is proposed to refine queries for new classes while preserving unrefined features, which enables Importance-based Knowledge Distillation for remembering the key features.
    \item Through extensive quantitative and qualitative evaluations on the ADE20K dataset, we demonstrate the state-of-the-art performance of our model in both Class Incremental Semantic and Panoptic Segmentation tasks.
\end{itemize}

\section{Related Work}
\label{sec:related_work}

\subsection{Semantic and Panoptic Segmentation}\label{sec:ss_ps}

Semantic segmentation aims to classify each pixel in an image, which focuses on the complete coverage of object(s) in each category. 
The pioneering work FCN~\cite{fcn} and U-Net~\cite{unet} introduces a pixel-wise classification paradigm by removing the fully connected layers. 
Since then, numerous approaches~\cite{deeplabv1,deeplabv2, deeplabv3,chen2018encoder,object17,segnet17,refinenet17,segmenter21} have been proposed following the paradigm, containing effective modules such as ASPP~\cite{deeplabv3} and pyramid structure~\cite{pspnet17}. 
In contrast, panoptic segmentation~\cite{panoptic} combines semantic and instance segmentation, requiring the classification of every pixel while distinguishing instances of the same category. 
These two tasks were addressed separately due to the limitation of the pixel-wise classification. 
However, the emergence of the mask classification paradigm~\cite{mask2former,maskformer,der21,mask-trans22} introduced a universal, transformer-based framework capable of solving multiple segmentation tasks simultaneously. 
MaskFormer~\cite{maskformer} pioneered this paradigm by combining class-agnostic mask proposal generation and mask classification, enabling both semantic and panoptic segmentation simultaneously. 
Building upon this, Mask2Former~\cite{mask2former} enhanced performance by incorporating multi-scale feature fusion and masked attention techniques. 
However, when learning new categories beyond the initial data, mask classification methods remain susceptible to \textit{catastrophic forgetting}.

\subsection{Continual Segmentation}
\label{sec:cs}

In Class Incremental Learning~(CIL), fine-tuning models on new data often results in a performance drop on old categories, \textit{i.e.}, \textit{catastrophic forgetting}~\cite{catastrophic99}. 
To address this, various methods~\cite{incremental22,plop21,mib20,modeling22,tackling21,recall21,continual21,rbc22} have been proposed, including retaining representative old samples~\cite{recall21,gss19,ccbo20,gradient17,ssul21,saving24,alife22,continual23}, compensation losses~\cite{cs2k24,foundation23,sdr21,ewf23}, and prompt representations~\cite{introducing23,learning22,incre23}. 
These advancements in CIL raise interest in more challenging segmentation tasks. 
Class Incremental Semantic Segmentation~(CISS) was first introduced in MiB~\cite{mib20}, reconstructing background regions to address \textit{background shifting} during incremental steps.
Since then, most methods~\cite{dkd22,rcil22} have employed techniques such as knowledge distillation~\cite{lwf17}, pseudo-labels from previous models~\cite{plop21,comformer23}, and background redefinition~\cite{mib20,background24} to mitigate \textit{background shifting}. 
However, these methods primarily rely on pixel-wise classification models like DeepLabV3~\cite{deeplabv3}. 
Therefore, CoMFormer~\cite{comformer23} first utilizes the mask classification segmentation model Mask2Former~\cite{mask2former} enabling solving both CISS and Class Incremental Panoptic Segmentation~(CIPS) tasks. 
ECLIPSE~\cite{eclipse24} introduces dynamic model structure expansion methods to extend learnable parameters for new classes and freezes the old parameters. 
CoMasTRe~\cite{comastre24} distills queries matched to old queries with high probability on old classes. 
BalConpas~\cite{balconpas24} balance the class proportion on selecting representative replay samples.
However, these methods often struggle to balance acquisition and preservation. 
Our approach mitigates this conflict by separating queries to decouple the two objectives.

\section{Preliminaries}
\label{sec:preliminaries}

\subsection{Problem Definition}\label{sec:definition}
In continual segmentation, the model is tasked with an incremental learning challenge over $T$ steps.
At each step $t\in\{1, \ldots, T\}$, the model is trained to recognize a unique set of classes $\bm{C}^{t}$, where the intersection across all sets from $i$ to $T$ is empty $\bigcap_{t=1}^{T}\bm{C}^{t}= \varnothing$ and their union forms the complete set of classes $\bigcup_{t=1}^{T}\bm{C}^{t}=\mathcal{\bm{C}}$.
In the current step $t$, the training dataset $\bm{D}_{train}^t$ consists of image-label pairs $(\bm{x}^{t},\bm{y}^{t})$, where $\bm{x}^{t}$ represents an image and $\bm{y}^{t}$ its corresponding segmentation label.
The labels $\bm{y}^{t}$ are available only for the classes $\bm{C}^{t}$ that need to be learned at this step, while labels for previously learned classes $\bm{C}^{1:t-1}$ and future classes $\bm{C}^{t+1:T}$ are inaccessible.
After completing training at this step, the model must perform segmentation across all classes learned up to that point, $\bm{C}^{1:t}$, thereby preventing catastrophic forgetting of the old classes $\bm{C}^{1:t-1}$ while effectively acquiring new ones $\bm{C}^{t}$.

\subsection{Revisiting Mask2Former and Pseudo-Labeling}\label{sec:psd}

The recent advanced universal segmentation network, Mask2Former~\cite{mask2former}, has favored the class incremental semantic and panoptic segmentation~(CISS and CIPS) tasks for its innovative proposal-based structure.
Mask2Former deviates from traditional pixel-wise classification pipelines by introducing a class-agnostic mask prediction and classification mechanism.
Specifically, the backbone $\bm{f}_b$ and the pixel decoder $\bm{f}_p$ generate multi-scale features that interact with $N$ learnable queries through multiple self-attention and cross-attention layers in the transformer decoder $\bm{f}_t$. 
These $N$ output queries $Q\in\mathbb{R}^{C_{Q}\times N}$ undergo further processing through two linear layers to form mask embeddings $\mathcal{E}_{mask}\in \mathbb{R}^{D\times N}$ and class embeddings $\mathcal{E}_{cls} \in\mathbb{R}^{{(C+1)}\times N}$.
The mask embeddings $\mathcal{E}_{mask}$ yield $N$ mask predictions $\bm{M}\in\mathbb{R}^{N\times H\times W}$ via dot product between each mask embedding and per-pixel embeddings $\mathcal{E}_{pixel}\in\mathbb{R}^{D\times H\times W}$ from $\bm{f}_p$, while $\mathcal{E}_{cls}$ accounts for the classification predictions of these masks.
Note that the additional channel in $\mathcal{E}_{cls}$ represents the \textit{no-obj} class, an auxiliary category utilized during training but excluded during inference. 

In the contexts of CISS and CIPS, where ground truth for old classes $\bm{C}^{1:t-1}$ are inaccessible, pseudo-labeling becomes crucial.
PLOP~\cite{plop21} introduces this strategy by generating pixel-level pseudo ground truths for $\bm{C}^{1:t-1}$, adapted to the mask-based structure by  CoMFormer~\cite{comformer23}, the first method based on Mask2Former for CISS and CIPS tasks.
This adaptation involves weighting the mask predictions $\bm{M}$ by the corresponding maximum class embedding score after the softmax process over the $C+1$ dimensions, \textit{i.e.}, $\max_{c=0}^{\lvert \bm{C}^{0:t-1} \rvert}(\text{softmax}_{c=0}^{|\bm{C}^{0:t-1}|}(\mathcal{E}_{cls}))$ to form $\bm{M}_\omega$, with associated categories $C_\omega = \arg\max_{c=0}^{|\bm{C}^{0:t-1}|}\mathcal{E}_{cls} \in \mathbb{R}^{N}$.
The pseudo-labeling process then labels the pixels outside the union of $\bm{y}^t$ with the category $c$ that achieves the maximum score across $\bm{M}_w$ and $\bm{C}_w$: 
\begin{equation}
    \tilde{M}_c(h,w) = 
    \begin{cases}
        1  & 
        \begin{aligned}
            \text{if}\quad c= &C_\omega[{\arg\max}_{n=1}^NM_w(n, h,w)] \\
            &\vee \max_{c\in C^{t}} y^t(h, w)=0, 
        \end{aligned}\\
        0 & \text{else }
    \end{cases}
\end{equation}
where $n$ indexes the $N$ proposals, and $h, w$ are the pixel coordinates.
This pseudo-labeling strategy has become foundational for CISS and CIPS methods~\cite{comformer23}. 

\section{Proposed Method}
\label{sec:method}

In our proposed Conflict Mitigation via Branched Optimization~(CoMBO), we introduce three distinct strategies: 
\textit{Half-Distillation-Half-Learning Strategy} for targeted knowledge retention, 
\textit{Importance-Based Knowledge Distillation} to prioritize crucial features, 
and \textit{Query Conflict Reduction} for efficient class integration.

\subsection{Query Conflict Reduction}\label{sec:qcr}

As mentioned in Sec.~\ref{sec:psd}, both class embeddings $\mathcal{E}_{cls}$ and mask predictions $\bm{M}$ are generated from the queries $\bm{Q}$. 
This interface becomes a critical juncture of conflicts between retaining knowledge of old classes and acquiring new classes manifest, which cannot break through the bottleneck of overall performance by simply finding the balance point. 
\begin{figure}
    \centering
    \includegraphics[width=0.95\linewidth]{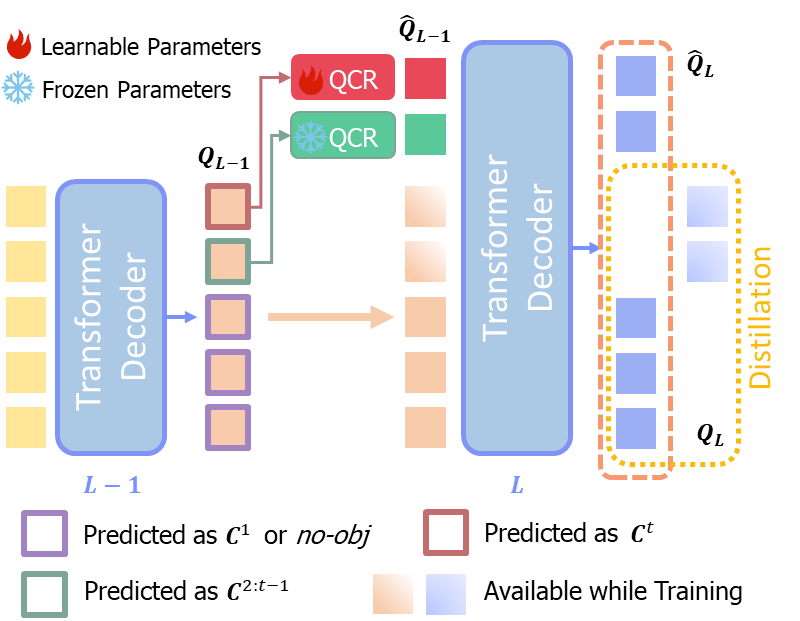}
    \caption{Illustration of the Query Conflict Reduction~(QCR) module. This module refines queries that predict incremental classes, allowing the processes of learning new features and retaining old features to occur separately.
    Note that the QCR module for previous incremental classes is frozen.}
    \label{fig:qcr}
\vspace{-0.3cm}
\end{figure}

To deconstruct the crossroad efficiently, we design a Query Conflict Reduction~(QCR) module $f_{QCR}(\cdot)$ as an additional adapter for new classes. 
As shown in Fig.~\ref{fig:qcr}, in most scenarios, queries from the previous layer $Q_{l-1}$ have the same classification results as the queries from the current layer due to the similar distribution and the same classification layer, especially in the latter queries. 
Due to the shared classifier and mask generator following different layers of the transformer decoder, the classification results on the same position of queries from adjacent layers are approximately the same.
Thus, we add the adapter for the queries from the penultimate layer $Q_{L-1}$ with classification results of new classes $\bm{C}^{2:t}$, as shown in Fig.~\ref{fig:qcr}. 
To be more specific, the queries $Q_{L-1, \tilde{c}}$ are selected for the adapter of incremental class $\tilde{c} \in \bm{C}^{2:t}$ via:
\begin{equation}
\resizebox{.9\linewidth}{!}{$
    Q_{L-1, \tilde{c}} = \{Q_{L-1}(n), n \in N \wedge \tilde{c}={\arg\max}_{c=0}^{|\bm{C}^{0:t}|}\mathcal{E}_{cls}(n)\},
    $}
\end{equation}
where $Q_{L-1, \tilde{c}}$ could contain none, one, or some selected queries that enable recognizing different instances. 
For each new category, we utilize the corresponding QCR module $f_{QCR,\tilde{c}}(\cdot)$ to refine the group of queries $Q_{L-1, \tilde{c}}$. 
Considering the efficiency and effectiveness, we introduce the low-rank two-layer adaptation as the QCR module: 
\begin{equation}
    \begin{aligned}
         \hat{Q}_{L-1, \tilde{c}} &=  f_{QCR,\tilde{c}}(Q_{L-1, \tilde{c}}) \\
         &=  Q_{L-1, \tilde{c}}W_1W_2 + Q_{L-1, \tilde{c}}, 
    \end{aligned}
\end{equation}
where $\hat{Q}_{L-1, \tilde{c}}$ represent the adapted queries of class $\tilde{c}$, $W_1\in \mathbb{R}^{D\times r}$ and $W_2\in \mathbb{R}^{r\times D}$ denotes the weights for QCR. 
The QCR module separates the refined queries $\hat{Q}_{L, \tilde{c}}$ from the original queries $Q_{L, \tilde{c}}$, enabling a bifurcate structure for simultaneously learning $\bm{C}^t$ with class-specific adaptation and keep memorizing the features of $Q_{L, \tilde{c}}$ as well as the logits of $\bm{C}^{0:t-1}$.

\subsection{Half-Distillation-Half-Learning Strategy}

Previous approaches based on Mask2Former usually separate learning and distillation processes on class embeddings $\mathcal{E}_{cls}$ due to its different learning strategies compared to pixel-wise segmentation structure~(\textit{e.g.} DeepLabv3~\cite{deeplabv3}), mentioned in Sec.~\ref{sec:psd}. 
This causes conflicts between distillation and learning, where the same logit could have contradictory optimization directions. 
Therefore, it is necessary to unify the optimization process towards a fixed target. 
\begin{figure}
    \centering
    \includegraphics[width=0.95\linewidth]{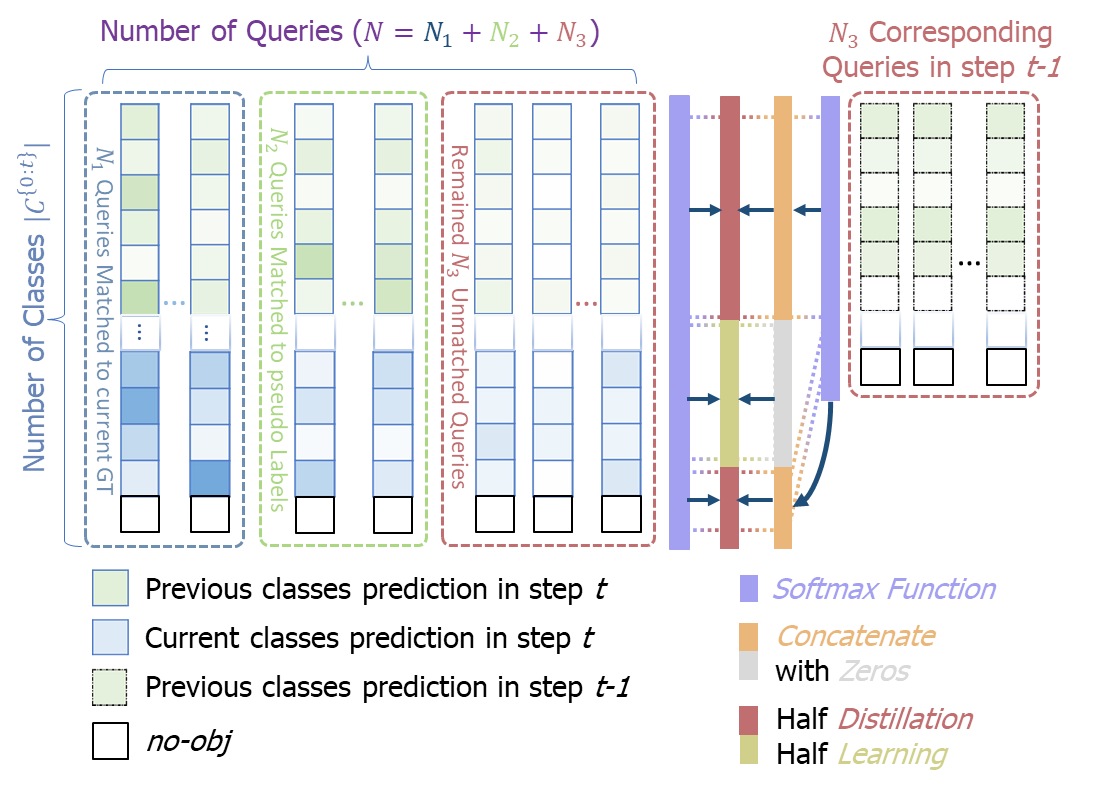}
    \caption{Details of the Half-Distillation-Half-Learning strategy on $N$ queries, including classification loss on matched queries and Kullback-Leibler Divergence loss on the other queries. 
    The latter involves the logits of both old and current classes. }
    \label{fig:hdhl}
    \vspace{-0.3cm}
\end{figure}

Our Half-Distillation-Half-Learning~(HDHL) strategy combines distillation and learning for $N$ class embeddings $\mathcal{E}_{cls}$, as illustrated in Fig.~\ref{fig:hdhl}. 
We follow the bipartite matching of Mask2Former and select embeddings $\mathcal{E}_{cls}^{\bm{C}^t}$ matched to labels of current categories $\bm{C}^t$ and $\mathcal{E}_{cls}^{\bm{C}^{1:t-1}}$ matched to pseudo labels of old categories $\bm{C}^{1:t-1}$.
These embeddings, following the original optimization strategy, are optimized by cross-entropy loss $\mathcal{L}_{cls}$. 
The unmatched embeddings $\mathcal{E}_{cls}^{\varnothing^t}$ are grouped under the no-object (\textit{no-obj}) category. 
However, adhering to the original learning strategy leads to \textit{catastrophic forgetting}. 
Besides, using the distillation strategy for old categories $\bm{C}^{1:t-1}$ using $\mathcal{E}_{cls}^{\varnothing^{t-1}}$ ignores the learning of current categories $\bm{C}^t$.
However, identifying regions that do not belong to $\bm{C}^t$ is just as important as learning those that do. 
Considering this, we design a strategy to simultaneously optimize the logits of $\bm{C}^{1:t-1}$ and $\bm{C}^t$ with a shared Kullback-Leibler Divergence $\mathcal{L}_{kl}$:
\begin{equation}
    \mathcal{L}_{kl} = \frac{1}{|\mathcal{E}_{cls}^{\varnothing^t}|} \sum\limits_{j=1}^{|\mathcal{E}_{cls}^{\varnothing^t}|} \sum\limits_{c=0}^{|\bm{C}^{0:t}|} \mathcal{\varphi}_{cls}^{\varnothing^{t-1}}(j, c) \log \frac{\mathcal{\varphi}_{cls}^{\varnothing^{t-1}}(j, c)}{\mathcal{\varphi}_{cls}^{\varnothing^t}(j, c)}, 
\end{equation}
where 
\begin{equation}
    \begin{aligned}
    \mathcal{\varphi}_{cls}^{\varnothing^t}(j, c) &= \mathcal{S}^t(j,c), \\
    \mathcal{\varphi}_{cls}^{\varnothing^{t-1}}(j, c) &= 
        \begin{cases}
          \mathcal{S}^{t-1}(j,c)  & \text{if}\  c\in \bm{C}^{0:t-1}, \\
            0 & \text{else},
        \end{cases}
    \end{aligned}
\end{equation}
and
\begin{equation}
    \mathcal{S}^{t}(j,c) =         \frac{\exp(\mathcal{E}_{cls}^{\varnothing^{t}}(j,c))}{ \sum_{\hat{c}=0}^{|\bm{C}^{0:t}|}\exp(\mathcal{E}_{cls}^{\varnothing^{t}}(j,\hat{c}))}
\end{equation}
represent the $\mathrm{softmax}$ activation. 

The $\mathcal{L}_{kl}$ not only distill the previous class embedding $\mathcal{E}_{cls}^{\varnothing^{t-1}}$ to the corresponding categories, including the \textit{no-obj} category, but decreasing the logits of current categories to figure out the objects apart from them. 
The unified HDHL strategy maintains the probability distribution of old logits seamlessly with the integration of new logits. 
By combining the original $\mathcal{L}_{cls}$ and the designed $\mathcal{L}_{kl}$ as $\mathcal{L}_{DL}$, the optimization process covers all the class embeddings without contradictory targets: 
\begin{equation}
    \mathcal{L}_{DL} = \lambda_{cls}\cdot \mathcal{L}_{cls} + \lambda_{kl}\cdot \mathcal{L}_{kl}, 
\end{equation}
where $\lambda_{cls}$ and $\lambda_{kl}$ control the balance between two losses. 

\subsection{Importance-Based Knowledge Distillation}
\label{sec:ikd}

Previous methods mitigate catastrophic forgetting by applying feature distillation, which may hinder the acquisition of new knowledge.
To address this, we propose an Importance-Based Knowledge Distillation~(IKD) method, in which stronger distillation is applied to features critical for old classes, while less emphasis is placed on less pertinent features, thereby balancing stability with plasticity. 

The importance measurement requires the cost matrix $\mathcal{A} \in \mathbb{R}^{N \times S}$, adopted in the bipartite matching process, where $S$ is the number of image labels. 
Values in $\mathcal{A}$ indicate the relationship between queries and classes.
For each image, the transformer decoder generates $N$ predicted segments, denoted as $\mathcal{Z} = \{(\mathcal{E}_{cls}(n), M(n))\}_{n=1}^{N}$, 
The cost matrix $\mathcal{A}$ is then computed based on the output segments $\mathcal{Z}$ and the image labels $\bm{y}^{t} = \{(c_{s}, m_{s})\}_{s=1}^{S}$:
\begin{equation}\label{eq:matrix}
     \mathcal{A}(n,s) = -\lambda_{cls} \cdot \mathcal{\varphi}_{cls}(n,c_{s}) + \lambda_{mask} \cdot \mathcal{L}_{mask}(\mathcal{E}_{mask}(n),m_{s}),
\end{equation}
where $\mathcal{\varphi}_{cls}$ is the classification output obtained by applying softmax to $\mathcal{E}_{cls}$, and $\lambda_{cls}$ and $\lambda_{mask}$ represent the corresponding hyperparameters.

After each training stage, we compute the cost matrix $\mathcal{A}$ across every image in the training set $\bm{D}^t_{train}$ of the current incremental step $t$. 
The minimum costs among the classes for $N$ segments are accumulated in the corresponding position of buffer $B^t \in \mathbb{R}^{N}$, where segments with lower minimum cost values are more likely to recognize and cover the regions of a class $\bm{C}^t$. 
By normalizing and reversing the values of $B^t$, we derive the importance matrix $I^t_{\bm{C}^t} \in \mathbb{R}^{N}$ of current classes $\bm{C}^t$.
The importance matrix $I^{t+1} \in \mathbb{R}^{N}$ is further estimated via the weighted accumulation of current importance matrix $I^t \in \mathbb{R}^{N}$ and $I^t_{\bm{C}^t}$, where the weights refer to the $|\bm{C}^{1:t-1}|$ and $|\bm{C}^t|$. 
The generated $I^{t+1}$ will be utilized in the next incremental training phase for weighted distillation of the queries $Q^t$, which is computed as follows:
\begin{equation}
     \mathcal{L}_{IKD} = \frac{1}{N}\sum_{n=1}^{N}I^{t}(n) \cdot \|Q^t(n), Q^{t-1}(n)\|_2^2,
\end{equation}
where $\|\cdot\|_2$ denotes the Euclidean distance, $Q^t(n)$ and $Q^{t-1}(n)$ denotes the transformer decoder features from the current model $f^{t}(\cdot)$ and the old model $f^{t-1}(\cdot)$, respectively. 
Algorithm~\ref{algorithm_1} presents the details of the importance estimation procedure after training $f^{t}(\cdot)$.

\begin{algorithm}
\caption{Importance Matrix Estimation}\label{algorithm_1}
\begin{algorithmic}[1]
\STATE \textbf{Input:} current model $f^{t}(\cdot)$, current dataset $\bm{D}^t_{train}$, current importance matrix $I^{t}$
\STATE \textbf{Initialization:} $I^{t+1} \leftarrow 0$ , $B^t \leftarrow 0$ 
\FOR{all $(\bm{x}^{t}, \bm{y}^{t}) \in \bm{D}^t_{train}$}
    \STATE Compute $\mathcal{A}$ following Eq.~(\ref{eq:matrix})
    \STATE Update the buffer:
    \STATE $B^t \leftarrow B^t + \min(\mathcal{A}, \mathrm{axis=1})$
\ENDFOR
\FOR{$n \in \{1,2,...,N\}$}
    \STATE $I^{t}_{C^t}(n) \leftarrow 1 - \frac{B^t(n) - \mathrm{min}(B^t)}{\mathrm{max}(B^t) - \mathrm{min}(B^t)}$
    \STATE $I^{t+1}(n) \leftarrow \frac{|\bm{C}^{1:t-1}|}{|\bm{C}^{1:t}|} \cdot I^{t}(n) + \frac{|\bm{C}^{t}|}{|\bm{C}^{1:t}|} \cdot I^{t}_{\bm{C}^t}(n)$
\ENDFOR
\STATE \textbf{Output:} Importance matrix $I^{t+1}$ for step $t+1$
\end{algorithmic}
\end{algorithm}

\subsection{Objective Function}

Overall, the objective function of CoMBO is defined as:
\begin{equation}
    \mathcal{L}_{H} = \mathcal{L}_{DL} + \lambda_{IKD}\mathcal{L}_{IKD}. 
\end{equation}
where $\mathcal{L}_{DL}$ represents the Half-Distillation-Half-Learning loss, and $\mathcal{L}_{IKD}$ means the Importance-Based Knowledge Distillation loss, with $\lambda_{IKD}$ as its weighting factor.
\begin{table*}[!htbp]
    \centering
    \resizebox{\textwidth}{!}{
    \begin{tabular}{r|ccc|ccc|ccc|ccc}
    \toprule
    \multirow{2}{*}{Method} & \multicolumn{3}{c|}{\textbf{100-50} (2 steps)} & \multicolumn{3}{c|}{\textbf{100-10} (6 steps)} & \multicolumn{3}{c|}{\textbf{100-5} (11 steps)} &  \multicolumn{3}{c}{\textbf{50-50}~(3 steps)}\\
    & 1-100 & 101-150 & all & 1-100 & 101-150 & all & 1-100 & 101-150 & all  & 1-50& 51-150&all\\
    \midrule
    FT & 0.0 & 1.3 & 0.4 & 0.0 & 2.9 & 1.0 & 0.0 & 25.8 & 8.6  & 0.0 & 12.0 &8.1 
\\
    MiB~\cite{mib20}\pub{CVPR20} & 35.1 & 19.3 & 29.8 & 27.1 & 10.0 & 21.4 & 24.0 & 6.5 & 18.1  & 42.4 & 15.5 &24.4 
\\
    PLOP~\cite{plop21}\pub{CVPR21} & 40.2 & 22.4 & 34.3 & 30.5 & 17.5 & 26.1 & 28.1 & 15.7 & 24.0  & 45.8 & 18.7 &27.7 
\\
    CoMFormer~\cite{comformer23}\pub{CVPR23} & 41.1 & \textbf{27.7} & 36.7 & 36.0 & 17.1 & 29.7 & 34.4 & 15.9 & 28.2  & 45.0 & 19.3 &27.9 
\\
    ECLIPSE~\cite{eclipse24}\pub{CVPR24} & 41.7 & 23.5 & 35.6 & 41.4 & 18.8 & 33.9 & 41.1 & 16.6 & \textbf{32.9}  & 46.0 & 20.7 &29.2 
\\
    BalConpas~\cite{balconpas24}\pub{ECCV24} & 42.8 & \underline{25.7} & \underline{37.1} & 40.7 & \underline{22.8} & \underline{34.7} & 36.1 & \underline{20.3} & 30.8  & 51.2 & \underline{26.5} &\underline{34.7} 
\\
    \midrule
    CoMBO~\pub{Ours} & 43.9 & 25.6 & \textbf{37.8} & 40.8 & \textbf{25.2} & \textbf{35.6} & 36.1 & \textbf{20.5} & \underline{30.9}  & 50.7 & \textbf{28.2} &\textbf{35.7} 
\\
    \midrule
    \rowcolor{gray!30}
    Joint & 43.8 & 30.9 & 39.5 & 43.8 & 30.9 & 39.5 & 43.8 & 30.9 & 39.5  & 50.7 & 33.9 &39.5 \\
    \bottomrule
    \end{tabular}
    }
    \caption{Quantitative comparison under Class Incremental Panoptic Segmentation with state-of-the-art exemplar-free methods on ADE20K in PQ. 
    Scores of novel classes and all classes in \textbf{bold} are the best while \underline{underlined} are the second best. }
    \label{tab:ade_ps}
\end{table*}

\section{Experimental Results} \label{sec:exp}
\subsection{Experimental Setup}
\paragraph{Datasets and Evaluation Metrics.}
We follow the experimental settings of previous works~\cite{comformer23,eclipse24,balconpas24} and evaluate our approach on the ADE20K~\cite{ade17} dataset.
The ADE20K dataset is specifically designed to support both panoptic and semantic segmentation tasks.
ADE20K contains 150 classes, including 100 \textit{thing} classes and 50 \textit{stuff} classes.
The dataset is composed of 20,210 images for training and 2,000 images for validation. 

For Class Incremental Semantic Segmentation~(CISS), we adopt the mean Intersection over Union~(mIoU) for evaluation. 
The Intersection over Union~(IoU) is calculated as $\text{IoU} = \frac{TP}{TP+FP+FN}$, where $TP$, $FP$, and $FN$ denote the number of true-positive, false-positive, and false-negative pixels, respectively. 
The mIoU metric averages the IoU across all classes for a more comprehensive evaluation.
For Class Incremental Panoptic Segmentation~(CIPS), following previous work~\cite{comformer23}, we employ Panoptic Quality~(PQ) as the evaluation metric.
PQ is defined as the product of Recognition Quality~(RQ) and Segmentation Quality~(SQ). 
To measure incremental learning capacity, we compute the corresponding metrics for the initial classes $\bm{C}^1$, incremental classes $\bm{C}^{2:T}$, and the aggregate of all classes $\bm{C}^{1:T}$. 

\paragraph{Protocols and Implementation Details.}
We follow previous incremental protocols and define scenarios as $N_{ini}-N_{inc}$, where $N_{ini}$ represents the number of initial classes, and $N_{inc}$ denotes the number of new classes introduced at each incremental step.
For example, in the $100\text{-}10$ scenario, the training begins with 100 classes, followed by the addition of 10 new classes per incremental step, without access to annotations of old classes. 
For both CIPS and CISS, we do evaluations on the following scenarios: $100\text{-}10$ (6 steps), $100\text{-}50$ (2 steps), $100\text{-}5$ (11 steps) and $50\text{-}50$ (3 steps). 

Our approach is based on the Mask2Former structure~\cite{mask2former}. 
Following previous approaches to CISS and CIPS, we utilize ResNet-50~\cite{resnet16} as the backbone for CIPS and ResNet-101 for CISS, with both pre-trained on ImageNet~\cite{imagenet}.
The input resolution of the images is $640\times640$ with an all-time batch size of 8.
We follow the training hyperparameter of Mask2Former in the initial step, with a learning rate of $10^{-4}$ and iterations of 160,000.
During the incremental steps, we set 1,000 iterations per class with a learning rate of $5\times10^{-5}$. 
The coefficients $r$, $\lambda_{cls}$, $\lambda_{kl}$, $\lambda_{IKD}$ are respectively set to 16, 2, 5, 3. 
All experiments were conducted on the NVIDIA RTX 4090. 

\begin{table*}[!htbp]
    \centering  
    \centering
    \resizebox{\textwidth}{!}{
    \begin{tabular}{r|ccc|ccc|ccc|ccc}
    \toprule
    \multirow{2}{*}{Method} & \multicolumn{3}{c|}{\textbf{100-50} (2 steps)} & \multicolumn{3}{c|}{\textbf{100-10} (6 steps)} & \multicolumn{3}{c|}{\textbf{100-5} (11 steps)} & \multicolumn{3}{c}{\textbf{50-50} (3 steps)}\\
    & 1-100 & 101-150 & all & 1-100 & 101-150 & all & 1-100 & 101-150 & all  & 1-50& 51-150&all\\
    \midrule
    FT & 0.0 & 26.7 & 8.9 & 0.0 & 2.3 & 0.8 & 0.0 & 1.1 & 0.3  & 0.0& 1.7&1.1\\
    MiB~\cite{mib20}\pub{CVPR20} & 37.0 & 24.1 & 32.6 & 23.5 & 10.6 & 26.6 & 21.0 & 6.1 & 16.1  & 45.6& 21.0&29.3\\
    PLOP~\cite{plop21}\pub{CVPR21} & 44.2 & 26.2 & 38.2 & 34.8 & 15.9 & 28.5 & 39.5 & 13.6 & 30.9  & 54.9& 30.2&38.4\\
    CoMFormer~\cite{comformer23}\pub{CVPR23} & 44.7 & 26.2 & 38.4 & 40.6 & 15.6 & 32.3 & 39.5 & 13.6 & 30.9  & 49.2& 26.6&34.1\\
    CoMasTRe~\cite{comastre24}\pub{CVPR24} & 45.7 & 26.0 & 39.2 & 42.3 & 18.4 & 34.4 & 40.8 & 15.8 & 32.6  & 49.8& 26.6&34.5\\
    ECLIPSE~\cite{eclipse24}\pub{CVPR24} & 45.0 & 21.7 & 37.1 & 43.4 & 17.4 & 34.6 & 43.3 & 16.3 & \underline{34.2}  & -& -&-\\
    BalConpas~\cite{balconpas24}\pub{ECCV24} & 49.9 & \underline{30.1} & \underline{43.3} & 47.3 & \underline{24.2} & \underline{38.6} & 42.1 & \underline{17.2} & 33.8  & 55.8& \underline{33.3} &\underline{40.8}\\
    \midrule
    CoMBO~\pub{Ours} & 50.2 & \textbf{34.4} & \textbf{44.9} & 47.8 & \textbf{27.7} & \textbf{41.1} & 44.6 & \textbf{22.6} & \textbf{37.3}  & 55.3& \textbf{36.9} & \textbf{43.0}\\
    \midrule
    \rowcolor{gray!30}
    Joint & 51.7 & 40.2 & 47.8 & 51.7 & 40.2 & 47.8 & 51.7 & 40.2 & 47.8  & 56.6& 43.5&47.8\\
    \bottomrule
    \end{tabular}  
    }
    \caption{Quantitative comparison under Class Incremental Semantic Segmentation with state-of-the-art exemplar-free methods on ADE20K in mIoU. 
    Scores of novel classes and all classes in \textbf{bold} are the best while \underline{underlined} are the second best. } 
    \label{tab:ade_ss}
\end{table*}

\subsection{Quantitative Results}
\paragraph{Comparisons in Class Incremental Panoptic Segmentation~(CIPS).}
We evaluated our approach against state-of-the-art exemplar-free methods on the ADE20K dataset within the CIPS framework, as detailed in Tab.~\ref{tab:ade_ps}. 
The performance of our CoMBO surpasses other methods in most of the scenarios, particularly in the more challenging 100-10 scenario, where it achieves a remarkable $35.6\%$ PQ with at least $3.4\%$ of advantage on incremental classes compared to the previous state-of-the-art methods.
This notable performance gain demonstrates the effectiveness of our CoMBO in reducing conflicts. 
Although our approach still underperforms the state-of-the-art approach ECLIPSE~\cite{eclipse24} in the 100-5 scenario, primarily due to its strong memorization ability from freezing parameters, our method achieves advanced performance for the new classes.
Moreover, in scenarios with fewer initial classes, \textit{i.e.}, 50-50 scenario, we maintain our advantage in learning new classes with a PQ of 28.2\% for incremental classes and 35.7\% for overall performance, widening the gap with strong distillation methods. 

\paragraph{Comparisons in Class Incremental Semantic Segmentation~(CISS).}
We further extended our evaluation to the semantic segmentation benchmark, comparing our approach with previous methods on the ADE20K dataset, as detailed in Tab.~\ref{tab:ade_ss}.
Our approach consistently outperforms previous methods in all tested scenarios. 
Notably, in the 100-10 scenario, our approach achieves a mIoU of $41.1\%$, surpassing the previous state-of-the-art by at least $3.5\%$ for incremental classes, demonstrating its effectiveness in mitigating conflicts.
Moreover, in the most challenging 100-5 long-term incremental scenario, our approach not only improves performance on old classes by $2.5\%$ but also achieves a significant $5.4\%$ increase in mIoU for new classes, ensuring both model stability and adaptability.

\section{Ablation Study}

\subsection{Component Ablations}\label{sec:ab_com}

We analyze the key components of our proposed framework, which includes the Half-Distillation-Half-Learning Strategy, Importance-based Knowledge Distillation, and the Query Conflict Reducing module. 
Our experiments focus on the 100-10 scenario in the CIPS, as it provides a moderate number of steps to examine the ability of both acquisition and retention. 
We evaluate various combinations of these components separately and present the results in Tab.~\ref{tab:ade_ablation}. 
The baseline method utilizes the vanilla loss with pseudo-labeling. 
\begin{table}[!htbp]
    \centering
    \begin{tabular}{c|c|c|ccc}
        \toprule
        \multirow{2}{*}{HDHL} & \multirow{2}{*}{IKD} & \multirow{2}{*}{QCR} & \multicolumn{3}{c}{\textbf{100-10} (6 steps)} \\ 
        &                       &                       & 1-100 & 101-150 & all \\ \midrule

        &                       &                       & 36.3  & 23.8    & 32.2  \\ 
        \checkmark&                       &             & 39.4  & 24.2     & 34.3   \\ 
        \checkmark&                       \checkmark&   & 40.0  & 24.6    & 34.9   \\ 
        &     \checkmark       &           \checkmark   & 38.8  & 25.4    & 34.4    \\ 
        \checkmark&           \checkmark&    \checkmark & 40.8  & 25.2    & \textbf{35.6}      \\ 
        \bottomrule
    \end{tabular} 
    \caption{Ablation study of the main components on the 100-10 task of CIPS. The baseline in the 1\textsuperscript{st} row employs vanilla losses with pseudo-labeling and excludes the QCR module.}
    \label{tab:ade_ablation}
    \vspace{-0.2cm}
\end{table}

Under identical experimental conditions, the HDHL Strategy significantly enhances the overall PQ by $2.1\%$, demonstrating its ability to unify logits of new classes into the previous logits distribution without contradictory losses. 
The implementation of the IKD mechanism has a remarkable $3.7\%$ increase in the PQ for old classes, underscoring its effectiveness in boosting the ability to retain the knowledge of old classes. 
Furthermore, when combined with the QCR module for branched optimization, CoMBO further boosts both the performance of old and new classes, achieving $0.8\%$ and $0.6\%$ gains in PQ, respectively, surpassing previous methods that struggle to balance acquisition and retention.
Consequently, CoMBO achieves a $3.4\%$ PQ improvement over the baseline. 
The above results demonstrate the effectiveness of the proposed CoMBO and its related modules, showing a noticeable improvement in reducing the conflict on model structures and losses. 

\subsection{Ablation Study of QCR}

\begin{figure}[!htbp]
    \centering
    \includegraphics[width=0.80\linewidth]{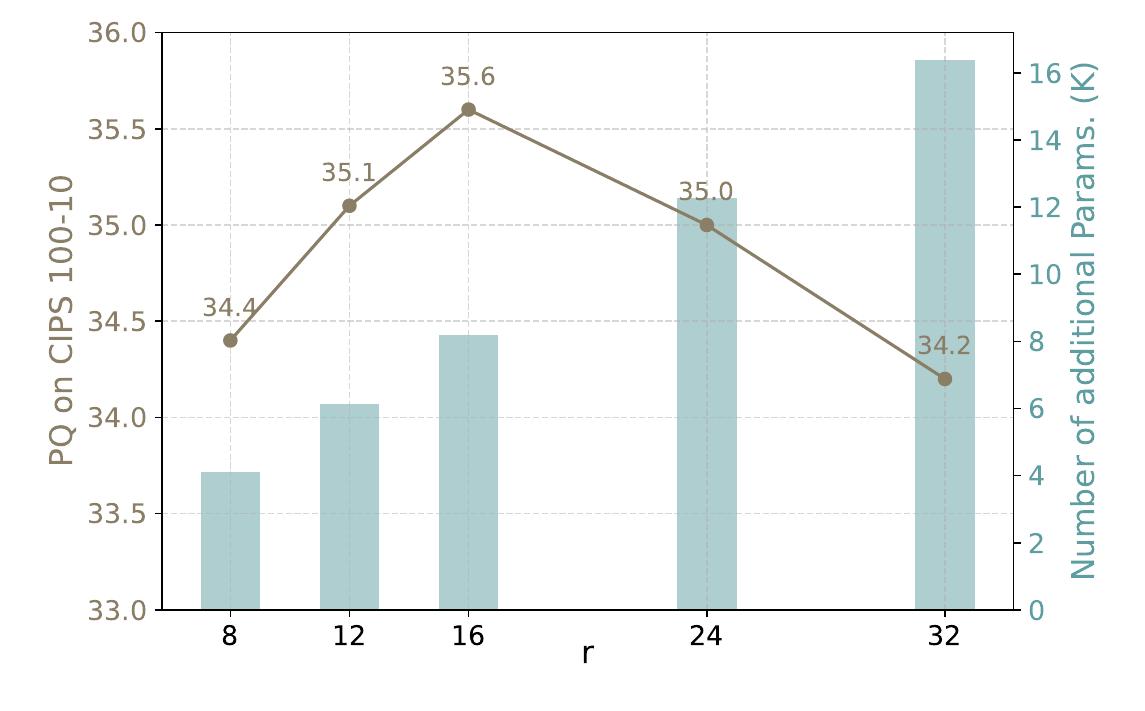}
    \vspace{-0.2cm}
    \caption{Ablation study of $r$ in QCR module, including PQ performance (left) and number of parameters (right).}
    \label{fig:rank}
    \vspace{-0.2cm}
\end{figure}

We validate the effectiveness of QCR with different settings of $r$ shown in Fig.~\ref{fig:rank}. 
Under the estimation of performance enhancement and additional parameters, our QCR module achieves the best result when $r=16$, where the quantity of additional parameters comes to 8.2K per class. 
Considering the total parameter of Mask2Former~\cite{mask2former} is 44.9M with ResNet-50, the additional parameters only reach up to 2\% of original parameters in total, which is 5 times less than per class additional parameters of ECLIPSE~\cite{eclipse24} while having better performance on new classes. 

\subsection{Ablation Study of HDHL Strategy }
\begin{figure*}[!htb]
    \centering    \includegraphics[width=0.82\linewidth]{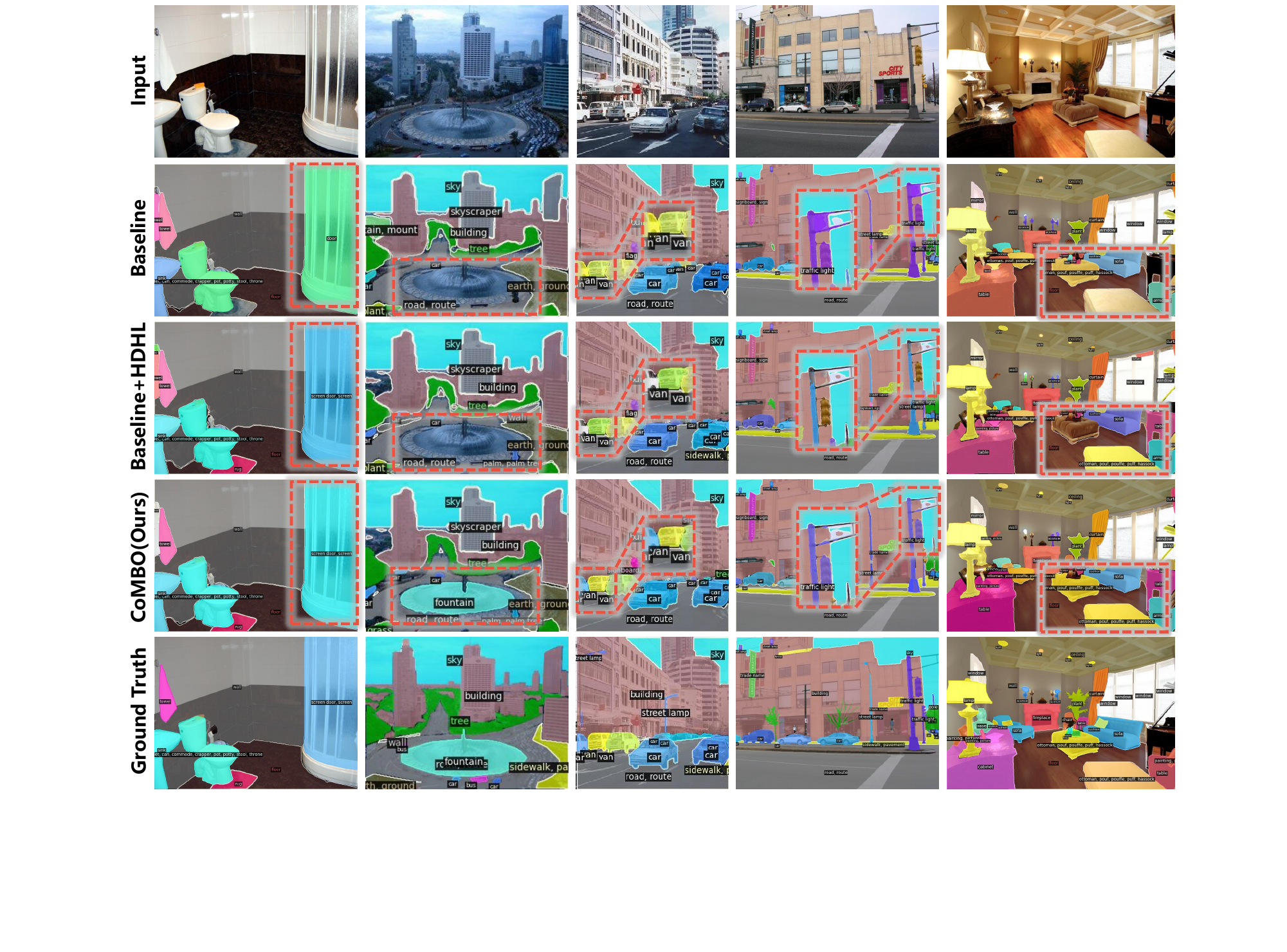}
    \vspace{-0.2cm}
    \caption{Qualitative results of CoMBO comparing to Baseline, Baseline+HDHL on 100-10 CIPS task of ADE20K. }
    \label{fig:qual}
    \vspace{-0.1cm}
\end{figure*}
To validate the effectiveness of our Half-Distillation-Half-Learning strategy, we conduct a series of experiments among other strategies, as shown in Tab.~\ref{tab:hdhl}. 
The 1\textsuperscript{st} row with $\mathcal{L}_{cls}$ on all classification embeddings represents the vanilla loss with pseudo-labeling, which is adopted in \cite{comformer23}. 
The 2\textsuperscript{nd} row utilizes an auxiliary distillation on the embeddings following~\cite{comastre24}. Compared to these strategies with tough supervision on unmatched embeddings $\mathcal{E}_{cls}^{t, \varnothing}$ and mostly conflicting losses, our HDHL strategy with soft and concentrated supervision has a better PQ of 34.3. 
Besides, the HDHL design of $\mathcal{L}_{kl}$ ensures the integrity while optimizing the unmatched embeddings $\mathcal{E}_{cls}^{t, \varnothing}$, with improvement of $2.1\%$ compared to distillation-only strategy in the 3\textsuperscript{rd} row and 0.8\% compared to the simple combination of distillation and $\mathcal{L}_{cls}$. 
The above results show that our HDHL strategy outperforms other learning or distillation strategies on the class embeddings. 

\begin{table}
    \centering
    \begin{tabular}{cc|cc|c}
    \toprule
         \multicolumn{2}{c|}{$\mathcal{L}_{cls}$} &\multicolumn{2}{c|}{$\mathcal{L}_{kl}$}&   \multirow{2}{*}{PQ}\\
         $\mathcal{E}_{cls}^{\bm{C}^t}$ \& $\mathcal{E}_{cls}^{\bm{C}^{1:t-1}}$ & $\mathcal{E}_{cls}^{t, \varnothing}$  & $\bm{C}^{0:t-1}$& $\bm{C}^{t}$&    \\
         \midrule
         $\checkmark$&  $\checkmark$ &  &  &  32.2\\
         $\checkmark$&  $\checkmark$ &  $\checkmark$&  &  32.6\\
         $\checkmark$&   &  $\checkmark$&  &  32.2\\
        $\checkmark$& $\checkmark$& $\checkmark$& & 33.5\\
        $\checkmark$&  & $\checkmark$& $\checkmark$& \textbf{34.3}\\
        \bottomrule
    \end{tabular}
    \caption{Ablation study of the HDHL strategy on task 100-10 of CIPS. 
    The 1\textsuperscript{st} row represents the vanilla loss with pseudo-labeling. The last row represents our $\mathcal{L}_{DL}$.}
    \label{tab:hdhl}
    \vspace{-0.3cm}
\end{table}

\subsection{Qualitative Analysis}

We conduct the qualitative analysis~(shown in Fig.~\ref{fig:qual}) by comparing our proposed CoMBO to the \textit{Baseline} and \textit{Baseline+HDHL}~(1\textsuperscript{st} and 2\textsuperscript{nd} rows in Tab.~\ref{tab:ade_ablation}). 
In the 1\textsuperscript{st}, 2\textsuperscript{nd}, and 5\textsuperscript{th} columns of Fig.~\ref{fig:qual}, our CoMBO can indicate the incremental classes, \textit{i.e.}, \textit{screen door}, \textit{fountain}, and \textit{ottoman}, while other methods fail to recognize all these objects. 
In the 3\textsuperscript{rd} column, our method successfully remembers the \textit{car} in the left instead of covering the object with the prediction of the incremental class \textit{van}. 
Furthermore, our method allows more precise mask prediction of new class \textit{Traffic Light} in the 4\textsuperscript{th} column, demonstrating the effectiveness of refinements on the queries corresponding to new classes. 

\section{Conclusions}

In this paper, we present CoMBO, a novel Class Incremental Segmentation~(CIS) method for mitigating the conflict between acquisition and retention losses. 
This conflict arises from competing goals of efficiently acquiring knowledge about new classes while preserving knowledge of previously learned ones. 
Our QCR module branches the conflicting optimization targets via lightweight class-specific adaptation on queries, enabling the coexistence of learning and distillation on separated queries. 
The HDHL strategy and IKD further reduce the contradictory optimization on classification logits and queries with unified targets, respectively, thereby enhancing the overall model performance. 
Extensive experimental validation on the ADE20K dataset underscores the superiority of our approach, demonstrating the effectiveness of each component in the CIS task, particularly excelling in performance on new classes.

\clearpage
\section*{Acknowledgments} This work was supported by the National Key R\&D Program of China (Grant 2022YFC3310200), the National Natural Science Foundation of China (Grant 62472033, 92470203), the Beijing Natural Science Foundation (Grant L242022), the Royal Society grants (SIF\textbackslash R1\textbackslash231009,  IES\textbackslash R3\textbackslash223050) and an Amazon Research Award.

{
    \small    \bibliographystyle{ieeenat_fullname}
    \bibliography{ref}

\begin{thebibliography}{52}
\providecommand{\natexlab}[1]{#1}
\providecommand{\url}[1]{\texttt{#1}}
\expandafter\ifx\csname urlstyle\endcsname\relax
  \providecommand{\doi}[1]{doi: #1}\else
  \providecommand{\doi}{doi: \begingroup \urlstyle{rm}\Url}\fi

\bibitem[Aljundi et~al.(2019)Aljundi, Lin, Goujaud, and Bengio]{gss19}
Rahaf Aljundi, Min Lin, Baptiste Goujaud, and Yoshua Bengio.
\newblock Gradient based sample selection for online continual learning.
\newblock \emph{{Adv. Neural Inf. Process. Syst.}}, 32, 2019.

\bibitem[Badrinarayanan et~al.(2017)Badrinarayanan, Kendall, and Cipolla]{segnet17}
Vijay Badrinarayanan, Alex Kendall, and Roberto Cipolla.
\newblock Segnet: A deep convolutional encoder-decoder architecture for image segmentation.
\newblock \emph{{IEEE Trans. on Pattern Anal. Mach. Intell.}}, 39\penalty0 (12):\penalty0 2481--2495, 2017.

\bibitem[Baek et~al.(2022)Baek, Oh, Lee, Lee, and Ham]{dkd22}
Donghyeon Baek, Youngmin Oh, Sanghoon Lee, Junghyup Lee, and Bumsub Ham.
\newblock Decomposed knowledge distillation for class-incremental semantic segmentation.
\newblock \emph{{Adv. Neural Inf. Process. Syst.}}, 35:\penalty0 10380--10392, 2022.

\bibitem[Borsos et~al.(2020)Borsos, Mutny, and Krause]{ccbo20}
Zal{\'a}n Borsos, Mojmir Mutny, and Andreas Krause.
\newblock Coresets via bilevel optimization for continual learning and streaming.
\newblock \emph{{Adv. Neural Inf. Process. Syst.}}, 33:\penalty0 14879--14890, 2020.

\bibitem[Cermelli et~al.(2020)Cermelli, Mancini, Bulo, Ricci, and Caputo]{mib20}
Fabio Cermelli, Massimiliano Mancini, Samuel~Rota Bulo, Elisa Ricci, and Barbara Caputo.
\newblock Modeling the background for incremental learning in semantic segmentation.
\newblock In \emph{{Proc. IEEE Conf. Comput. Vis. Pattern Recognit.}}, pages 9233--9242, 2020.

\bibitem[Cermelli et~al.(2022{\natexlab{a}})Cermelli, Fontanel, Tavera, Ciccone, and Caputo]{incremental22}
Fabio Cermelli, Dario Fontanel, Antonio Tavera, Marco Ciccone, and Barbara Caputo.
\newblock Incremental learning in semantic segmentation from image labels.
\newblock In \emph{{Proc. IEEE Conf. Comput. Vis. Pattern Recognit.}}, pages 4371--4381, 2022{\natexlab{a}}.

\bibitem[Cermelli et~al.(2022{\natexlab{b}})Cermelli, Geraci, Fontanel, and Caputo]{modeling22}
Fabio Cermelli, Antonino Geraci, Dario Fontanel, and Barbara Caputo.
\newblock Modeling missing annotations for incremental learning in object detection.
\newblock In \emph{{Proc. IEEE Conf. Comput. Vis. Pattern Recognit.}}, pages 3700--3710, 2022{\natexlab{b}}.

\bibitem[Cermelli et~al.(2023)Cermelli, Cord, and Douillard]{comformer23}
Fabio Cermelli, Matthieu Cord, and Arthur Douillard.
\newblock Comformer: Continual learning in semantic and panoptic segmentation.
\newblock In \emph{{Proc. IEEE Conf. Comput. Vis. Pattern Recognit.}}, pages 3010--3020, 2023.

\bibitem[Cha et~al.(2021)Cha, Yoo, Moon, et~al.]{ssul21}
Sungmin Cha, YoungJoon Yoo, Taesup Moon, et~al.
\newblock Ssul: Semantic segmentation with unknown label for exemplar-based class-incremental learning.
\newblock \emph{{Adv. Neural Inf. Process. Syst.}}, 34:\penalty0 10919--10930, 2021.

\bibitem[Chen et~al.(2024)Chen, Cong, Luo, Ip, and Kwong]{saving24}
Jinpeng Chen, Runmin Cong, Yuxuan Luo, Horace Ip, and Sam Kwong.
\newblock Saving 100x storage: prototype replay for reconstructing training sample distribution in class-incremental semantic segmentation.
\newblock \emph{{Adv. Neural Inf. Process. Syst.}}, 36, 2024.

\bibitem[Chen et~al.(2025)Chen, Cong, Luo, Ip, and Kwong]{balconpas24}
Jinpeng Chen, Runmin Cong, Yuxuan Luo, Horace Ho~Shing Ip, and Sam Kwong.
\newblock Strike a balance in continual panoptic segmentation.
\newblock In \emph{{Proc. Eur. Conf. Comput. Vis.}}, pages 126--142. Springer, 2025.

\bibitem[Chen(2014)]{deeplabv1}
Liang-Chieh Chen.
\newblock Semantic image segmentation with deep convolutional nets and fully connected crfs.
\newblock \emph{arXiv preprint arXiv:1412.7062}, 2014.

\bibitem[Chen(2017)]{deeplabv3}
L.~C. Chen.
\newblock Rethinking atrous convolution for semantic image segmentation.
\newblock \emph{arXiv preprint:1706.05587}, 2017.

\bibitem[Chen et~al.(2017)Chen, Papandreou, and et~al.]{deeplabv2}
Liang-Chieh Chen, George Papandreou, and et al.
\newblock Deeplab: Semantic image segmentation with deep convolutional nets, atrous convolution, and fully connected crfs.
\newblock \emph{{IEEE Trans. on Pattern Anal. Mach. Intell.}}, 40\penalty0 (4):\penalty0 834--848, 2017.

\bibitem[Chen et~al.(2018)Chen, Zhu, Papandreou, Schroff, and Adam]{chen2018encoder}
Liang-Chieh Chen, Yukun Zhu, George Papandreou, Florian Schroff, and Hartwig Adam.
\newblock Encoder-decoder with atrous separable convolution for semantic image segmentation.
\newblock In \emph{{Proc. Eur. Conf. Comput. Vis.}}, pages 801--818, 2018.

\bibitem[Cheng et~al.(2021)Cheng, Schwing, and Kirillov]{maskformer}
Bowen Cheng, Alex Schwing, and Alexander Kirillov.
\newblock Per-pixel classification is not all you need for semantic segmentation.
\newblock \emph{{Adv. Neural Inf. Process. Syst.}}, 34:\penalty0 17864--17875, 2021.

\bibitem[Cheng et~al.(2022)Cheng, Misra, Schwing, Kirillov, and Girdhar]{mask2former}
Bowen Cheng, Ishan Misra, Alexander~G Schwing, Alexander Kirillov, and Rohit Girdhar.
\newblock Masked-attention mask transformer for universal image segmentation.
\newblock In \emph{{Proc. IEEE Conf. Comput. Vis. Pattern Recognit.}}, pages 1290--1299, 2022.

\bibitem[Cong et~al.(2025)Cong, Cong, Liu, and Sun]{cs2k24}
Wei Cong, Yang Cong, Yuyang Liu, and Gan Sun.
\newblock Cs2k: Class-specific and class-shared knowledge guidance for incremental semantic segmentation.
\newblock In \emph{{Proc. Eur. Conf. Comput. Vis.}}, pages 244--261, 2025.

\bibitem[Deng et~al.(2009)Deng, Dong, Socher, Li, Li, and Fei-Fei]{imagenet}
Jia Deng, Wei Dong, Richard Socher, Li-Jia Li, Kai Li, and Li Fei-Fei.
\newblock Imagenet: A large-scale hierarchical image database.
\newblock In \emph{{Proc. IEEE Conf. Comput. Vis. Pattern Recognit.}}, pages 248--255, 2009.

\bibitem[Douillard et~al.(2021{\natexlab{a}})Douillard, Chen, Dapogny, and Cord]{plop21}
Arthur Douillard, Yifu Chen, Arnaud Dapogny, and Matthieu Cord.
\newblock Plop: Learning without forgetting for continual semantic segmentation.
\newblock In \emph{{Proc. IEEE Conf. Comput. Vis. Pattern Recognit.}}, pages 4040--4050, 2021{\natexlab{a}}.

\bibitem[Douillard et~al.(2021{\natexlab{b}})Douillard, Chen, Dapogny, and Cord]{tackling21}
Arthur Douillard, Yifu Chen, Arnaud Dapogny, and Matthieu Cord.
\newblock Tackling catastrophic forgetting and background shift in continual semantic segmentation.
\newblock \emph{arXiv preprint arXiv:2106.15287}, 2021{\natexlab{b}}.

\bibitem[French(1999)]{catastrophic99}
Robert~M French.
\newblock Catastrophic forgetting in connectionist networks.
\newblock \emph{Trends in cognitive sciences}, 3\penalty0 (4):\penalty0 128--135, 1999.

\bibitem[Gong et~al.(2024)Gong, Yu, Wang, and Xiao]{comastre24}
Yizheng Gong, Siyue Yu, Xiaoyang Wang, and Jimin Xiao.
\newblock Continual segmentation with disentangled objectness learning and class recognition.
\newblock In \emph{{Proc. IEEE Conf. Comput. Vis. Pattern Recognit.}}, pages 3848--3857, 2024.

\bibitem[He et~al.(2016)He, Zhang, Ren, and Sun]{resnet16}
Kaiming He, Xiangyu Zhang, Shaoqing Ren, and Jian Sun.
\newblock Deep residual learning for image recognition.
\newblock In \emph{{Proc. IEEE Conf. Comput. Vis. Pattern Recognit.}}, pages 770--778, 2016.

\bibitem[Khan et~al.(2023)Khan, Naeem, Van~Gool, Stricker, Tombari, and Afzal]{introducing23}
M.~G. Z.~A. Khan, M.~F. Naeem, Luc Van~Gool, Didier Stricker, Federico Tombari, and Muhammad~Zeshan Afzal.
\newblock Introducing language guidance in prompt-based continual learning.
\newblock In \emph{{Proc. IEEE Int. Conf. Comput. Vis.}}, pages 11463--11473, 2023.

\bibitem[Kim et~al.(2024)Kim, Yu, and Hwang]{eclipse24}
Beomyoung Kim, Joonsang Yu, and Sung~Ju Hwang.
\newblock Eclipse: Efficient continual learning in panoptic segmentation with visual prompt tuning.
\newblock In \emph{{Proc. IEEE Conf. Comput. Vis. Pattern Recognit.}}, pages 3346--3356, 2024.

\bibitem[Kirillov et~al.(2019)Kirillov, He, Girshick, Rother, and Doll{\'a}r]{panoptic}
Alexander Kirillov, Kaiming He, Ross Girshick, Carsten Rother, and Piotr Doll{\'a}r.
\newblock Panoptic segmentation.
\newblock In \emph{{Proc. IEEE Conf. Comput. Vis. Pattern Recognit.}}, pages 9404--9413, 2019.

\bibitem[Li and Hoiem(2017)]{lwf17}
Zhizhong Li and Derek Hoiem.
\newblock Learning without forgetting.
\newblock \emph{{IEEE Trans. on Pattern Anal. Mach. Intell.}}, 40\penalty0 (12):\penalty0 2935--2947, 2017.

\bibitem[Lin et~al.(2017)Lin, Milan, Shen, and Reid]{refinenet17}
Guosheng Lin, Anton Milan, Chunhua Shen, and Ian Reid.
\newblock Refinenet: Multi-path refinement networks for high-resolution semantic segmentation.
\newblock In \emph{{Proc. IEEE Conf. Comput. Vis. Pattern Recognit.}}, pages 1925--1934, 2017.

\bibitem[Long et~al.(2015)Long, Shelhamer, and Darrell]{fcn}
Jonathan Long, Evan Shelhamer, and Trevor Darrell.
\newblock Fully convolutional networks for semantic segmentation.
\newblock In \emph{{Proc. IEEE Conf. Comput. Vis. Pattern Recognit.}}, pages 3431--3440, 2015.

\bibitem[Lopez-Paz and Ranzato(2017)]{gradient17}
David Lopez-Paz and Marc'Aurelio Ranzato.
\newblock Gradient episodic memory for continual learning.
\newblock \emph{{Adv. Neural Inf. Process. Syst.}}, 30, 2017.

\bibitem[Maracani et~al.(2021)Maracani, Michieli, Toldo, and Zanuttigh]{recall21}
Andrea Maracani, Umberto Michieli, Marco Toldo, and Pietro Zanuttigh.
\newblock Recall: Replay-based continual learning in semantic segmentation.
\newblock In \emph{{Proc. IEEE Int. Conf. Comput. Vis.}}, pages 7026--7035, 2021.

\bibitem[Michieli and Zanuttigh(2021{\natexlab{a}})]{continual21}
Umberto Michieli and Pietro Zanuttigh.
\newblock Continual semantic segmentation via repulsion-attraction of sparse and disentangled latent representations.
\newblock In \emph{{Proc. IEEE Conf. Comput. Vis. Pattern Recognit.}}, pages 1114--1124, 2021{\natexlab{a}}.

\bibitem[Michieli and Zanuttigh(2021{\natexlab{b}})]{sdr21}
Umberto Michieli and Pietro Zanuttigh.
\newblock Continual semantic segmentation via repulsion-attraction of sparse and disentangled latent representations.
\newblock In \emph{{Proc. IEEE Conf. Comput. Vis. Pattern Recognit.}}, pages 1114--1124, 2021{\natexlab{b}}.

\bibitem[Oh et~al.(2022)Oh, Baek, and Ham]{alife22}
Youngmin Oh, Donghyeon Baek, and Bumsub Ham.
\newblock Alife: Adaptive logit regularizer and feature replay for incremental semantic segmentation.
\newblock \emph{{Adv. Neural Inf. Process. Syst.}}, 35:\penalty0 14516--14528, 2022.

\bibitem[Ronneberger et~al.(2015)Ronneberger, Fischer, and Brox]{unet}
Olaf Ronneberger, Philipp Fischer, and Thomas Brox.
\newblock U-net: Convolutional networks for biomedical image segmentation.
\newblock In \emph{{Proc. International Conference on Medical Image Computing and Computer-Assisted Intervention}}, pages 234--241, 2015.

\bibitem[Shang et~al.(2023)Shang, Li, Meng, Wu, Qiu, and Wang]{incre23}
Chao Shang, Hongliang Li, Fanman Meng, Qingbo Wu, Heqian Qiu, and Lanxiao Wang.
\newblock Incrementer: Transformer for class-incremental semantic segmentation with knowledge distillation focusing on old class.
\newblock In \emph{{Proc. IEEE Conf. Comput. Vis. Pattern Recognit.}}, pages 7214--7224, 2023.

\bibitem[Strudel et~al.(2021)Strudel, Garcia, Laptev, and Schmid]{segmenter21}
Robin Strudel, Ricardo Garcia, Ivan Laptev, and Cordelia Schmid.
\newblock Segmenter: Transformer for semantic segmentation.
\newblock In \emph{{Proc. IEEE Int. Conf. Comput. Vis.}}, pages 7262--7272, 2021.

\bibitem[Wang et~al.(2022)Wang, Zhang, Lee, Zhang, Sun, Ren, Su, Perot, Dy, and Pfister]{learning22}
Zifeng Wang, Zizhao Zhang, Chen-Yu Lee, Han Zhang, Ruoxi Sun, Xiaoqi Ren, Guolong Su, Vincent Perot, Jennifer Dy, and Tomas Pfister.
\newblock Learning to prompt for continual learning.
\newblock In \emph{{Proc. IEEE Conf. Comput. Vis. Pattern Recognit.}}, pages 139--149, 2022.

\bibitem[Wei et~al.(2017)Wei, Feng, Liang, Cheng, Zhao, and Yan]{object17}
Yunchao Wei, Jiashi Feng, Xiaodan Liang, Ming-Ming Cheng, Yao Zhao, and Shuicheng Yan.
\newblock Object region mining with adversarial erasing: A simple classification to semantic segmentation approach.
\newblock In \emph{{Proc. IEEE Conf. Comput. Vis. Pattern Recognit.}}, pages 1568--1576, 2017.

\bibitem[Xiao et~al.(2023)Xiao, Zhang, Feng, Liu, van~de Weijer, and Cheng]{ewf23}
Jia-Wen Xiao, Chang-Bin Zhang, Jiekang Feng, Xialei Liu, Joost van~de Weijer, and Ming-Ming Cheng.
\newblock Endpoints weight fusion for class incremental semantic segmentation.
\newblock In \emph{{Proc. IEEE Conf. Comput. Vis. Pattern Recognit.}}, pages 7204--7213, 2023.

\bibitem[Yan et~al.(2021)Yan, Xie, and He]{der21}
Shipeng Yan, Jiangwei Xie, and Xuming He.
\newblock Der: Dynamically expandable representation for class incremental learning.
\newblock In \emph{{Proc. IEEE Conf. Comput. Vis. Pattern Recognit.}}, pages 3014--3023, 2021.

\bibitem[Yu et~al.(2023)Yu, Zhou, Li, Yuan, Wang, and Wang]{foundation23}
Chaohui Yu, Qiang Zhou, Jingliang Li, Jianlong Yuan, Zhibin Wang, and Fan Wang.
\newblock Foundation model drives weakly incremental learning for semantic segmentation.
\newblock In \emph{{Proc. IEEE Conf. Comput. Vis. Pattern Recognit.}}, pages 23685--23694, 2023.

\bibitem[Yu et~al.(2022)Yu, Wang, Qiao, Collins, Zhu, Adam, Yuille, and Chen]{mask-trans22}
Qihang Yu, Huiyu Wang, Siyuan Qiao, Maxwell Collins, Yukun Zhu, Hartwig Adam, Alan Yuille, and Liang-Chieh Chen.
\newblock k-means mask transformer.
\newblock In \emph{{Proc. Eur. Conf. Comput. Vis.}}, pages 288--307, 2022.

\bibitem[Zhang and Gao(2024)]{background24}
Anqi Zhang and Guangyu Gao.
\newblock Background adaptation with residual modeling for exemplar-free class-incremental semantic segmentation.
\newblock In \emph{{Proc. Eur. Conf. Comput. Vis.}}, 2024.

\bibitem[Zhang et~al.(2022{\natexlab{a}})Zhang, Xiao, Liu, Chen, and Cheng]{rcil22}
Chang-Bin Zhang, Jia-Wen Xiao, Xialei Liu, Ying-Cong Chen, and Ming-Ming Cheng.
\newblock Representation compensation networks for continual semantic segmentation.
\newblock In \emph{{Proc. IEEE Conf. Comput. Vis. Pattern Recognit.}}, pages 7053--7064, 2022{\natexlab{a}}.

\bibitem[Zhang et~al.(2022{\natexlab{b}})Zhang, Gao, Fang, Jiao, and Wei]{micro22}
Zekang Zhang, Guangyu Gao, Zhiyuan Fang, Jianbo Jiao, and Yunchao Wei.
\newblock Mining unseen classes via regional objectness: A simple baseline for incremental segmentation.
\newblock \emph{{Adv. Neural Inf. Process. Syst.}}, 35:\penalty0 24340--24353, 2022{\natexlab{b}}.

\bibitem[Zhang et~al.(2023)Zhang, Gao, Jiao, Liu, and Wei]{coinseg23}
Zekang Zhang, Guangyu Gao, Jianbo Jiao, Chi~Harold Liu, and Yunchao Wei.
\newblock Coinseg: Contrast inter-and intra-class representations for incremental segmentation.
\newblock In \emph{{Proc. IEEE Int. Conf. Comput. Vis.}}, pages 843--853, 2023.

\bibitem[Zhao et~al.(2017)Zhao, Shi, Qi, Wang, and Jia]{pspnet17}
Hengshuang Zhao, Jianping Shi, Xiaojuan Qi, Xiaogang Wang, and Jiaya Jia.
\newblock Pyramid scene parsing network.
\newblock In \emph{{Proc. IEEE Conf. Comput. Vis. Pattern Recognit.}}, pages 2881--2890, 2017.

\bibitem[Zhao et~al.(2022)Zhao, Yang, Fu, and Li]{rbc22}
Hanbin Zhao, Fengyu Yang, Xinghe Fu, and Xi Li.
\newblock Rbc: Rectifying the biased context in continual semantic segmentation.
\newblock In \emph{{Proc. Eur. Conf. Comput. Vis.}}, pages 55--72, 2022.

\bibitem[Zhou et~al.(2017)Zhou, Zhao, Puig, Fidler, Barriuso, and Torralba]{ade17}
Bolei Zhou, Hang Zhao, Xavier Puig, Sanja Fidler, Adela Barriuso, and Antonio Torralba.
\newblock Scene parsing through ade20k dataset.
\newblock In \emph{{Proc. IEEE Conf. Comput. Vis. Pattern Recognit.}}, pages 633--641, 2017.

\bibitem[Zhu et~al.(2023)Zhu, Chen, Yin, See, and Liu]{continual23}
Lanyun Zhu, Tianrun Chen, Jianxiong Yin, Simon See, and Jun Liu.
\newblock Continual semantic segmentation with automatic memory sample selection.
\newblock In \emph{{Proc. IEEE Conf. Comput. Vis. Pattern Recognit.}}, pages 3082--3092, 2023.

\end{thebibliography}
}

\clearpage
\setcounter{page}{1}
\maketitlesupplementary

\begin{center}
  \large \textbf{Contents}
\end{center}
\startcontents 
\printcontents{}{1}{} 
\section{More Analysis}
\label{sec:moreanalysis}
In this section, we present additional experimental results to further examine the effectiveness of the proposed components, including an ablation study of parameter freezing on queries and QCR modules, component ablations within CISS, an ablation study of IKD, and an analysis of prediction results from different layers of class embeddings. These experiments provide us with a deeper understanding of the individual contributions and interactions of each component, shedding light on their specific roles and the ways in which they enhance overall system performance. 
\subsection{Ablation Study of Parameter Freezing}

As mentioned in Fig.~\ref{fig:qcr}, the parameters of the query embeddings $Q$ and the QCR modules corresponding to the previous incremental classes $C^{2:t-1}$ are frozen during step $t$. 
To evaluate the effectiveness of this parameter-freezing strategy in incremental learning, we conducted a series of experiments. 
As shown in Tab.~\ref{tab:ab_fz}, freezing both the query embeddings $Q$ and QCR modules $f_{QCR,\tilde{c}}$ with $\tilde{c}\in C^{2:t-1}$ results in an improvement of 0.4\% compared to the unfreezing method.
This demonstrates the strategy's efficacy in retaining knowledge of old classes while accommodating new classes. 
Thus, the results prove that the parameter freezing strategy avoids disturbing the impressionable query embeddings.
Besides, it is important to note that freezing the parameters of queries does not mean keeping the queries of $Q_{l}$ static when $l>1$. 
The optimization of $Q_{l}$ mainly affects features from the pixel-decoder, where the model integrates knowledge of new classes to enhance feature extraction.
Furthermore, the QCR module $f_{QCR, \tilde{c}}$, as a lightweight adapter, encounters challenges similar to query embeddings, where limited pseudo labels of previous incremental classes $C^{2:t-1}$ could result in overfitting and misguidance, causing 1.3\% decreasing on these incremental classes. 
In such cases, freezing the relevant parameters offers a simpler and more effective alternative to distillation, mitigating these risks and maintaining performance stability.

\begin{table}[!ht]
    
    \centering
    \begin{tabular}{c|c|ccc}
        \toprule
         \multirow{2}{*}{$Q$}& \multirow{2}{*}{$f_{QCR}$}& \multicolumn{3}{c}{\textbf{100-10} (6 steps)} \\ 
                               &                       & 1-100 & 101-150 & all \\ \midrule

                               \checkmark&                       \checkmark & 40.3& 25.2& 35.2\\ 
                               \checkmark&   & 40.7& 25.0& 35.5\\ 
             &           \checkmark   & 41.0& 23.9& 35.3\\ 
                   &    & 40.8  & 25.2    & \textbf{35.6}      \\ 
        \bottomrule
    \end{tabular} 
    \caption{Ablation study of the parameter freezing strategy on the query embeddings $Q$ and QCR modules $f_{QCR, \tilde{c}}$ of previous incremental classes $\tilde{c}\in C^{2:t-1}$. 
    The experiments are conducted on the CIPS 100-10 task of the ADE20K. Note that the ``$\checkmark$" denotes learnable parameters. }
    \label{tab:ab_fz}
\end{table}

\subsection{Component Ablations in CISS}\label{sec:sp_ab_com}

In this section, we evaluate the components of our proposed framework, including the Half-Distillation-Half-Learning Strategy, Importance-based Knowledge Distillation~(IKD), and Query Conflict Reducing~(QCR) module, in the 100-10 scenario of the CISS task.
We analyze various combinations of these components and present the results in Tab.~\ref{tab:ade_ablation_ciss}. 
The baseline approach employs standard losses from Mask2Former~\cite{mask2former} with pseudo-labeling.
Under the same experimental setup, the inclusion of the HDHL strategy leads to a significant improvement in overall mIoU by 3.4\%, highlighting its ability to seamlessly integrate logits from new classes into the existing logits distribution while avoiding conflicting losses. 
The introduction of IKD leads to an impressive 4.1\% increase in mIoU for old classes, showcasing its effectiveness in mitigating catastrophic forgetting by selectively distilling important knowledge. 
Additionally, by incorporating the QCR module for branched optimization, the proposed CoMBO further enhances performance on both old and new classes, with respective mIoU improvements of 0.8\% and 0.2\%, surpassing the limitations of previous state-of-the-art methods in balancing these two aspects. 
As a result, the overall mIoU increases by 4.6\% compared to the baseline. 
These findings emphasize the efficacy of the proposed approach CoMBO and its associated components in reducing conflicts within model structures and losses, achieving substantial performance gains.

\begin{table}[!ht]
    \centering
    \begin{tabular}{c|c|c|ccc}
        \toprule
        \multirow{2}{*}{HDHL} & \multirow{2}{*}{IKD} & \multirow{2}{*}{QCR} & \multicolumn{3}{c}{\textbf{100-10} (6 steps)} \\ 
        &                       &                       & 1-100 & 101-150 & all \\ \midrule

        &                       &                       & 42.9  & 23.6    & 36.5  \\ 
        \checkmark&                       &             &  46.5 &   26.8   &  39.9  \\ 
        \checkmark&                       \checkmark&   & 47.0  & 27.5   & 40.5   \\ 
        &     \checkmark       &           \checkmark   & 46.1  & 27.1   & 39.8    \\ 
        \checkmark&           \checkmark&    \checkmark & 47.8  & 27.7    & \textbf{41.1}      \\ 
        \bottomrule
    \end{tabular} 
    \caption{Ablation study of the main components on task 100-10 of CISS. 
    Baseline in the 1\textsuperscript{st} row uses vanilla losses with pseudo-labeling.} 
    \label{tab:ade_ablation_ciss}
\end{table}

\begin{figure}[!t]
    \centering
    \includegraphics[width=0.75\linewidth]{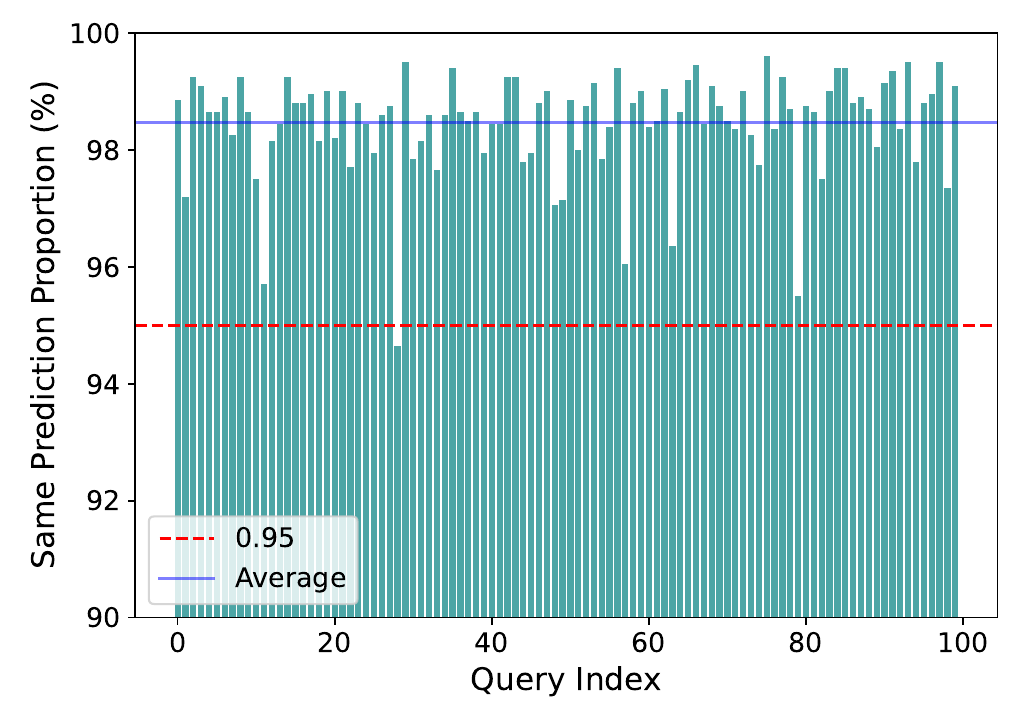}
    \caption{Proportion of samples with the same classification predictions between $\mathcal{E}_{cls, L-1}$ at layer $L-1$ and $\mathcal{E}_{cls,L}$ at layer $L$ without the QCR module. 
    The results indicate that nearly all embeddings from the queries have more than 95\% samples with the same predictions, with an average proportion exceeding 98\%.}
    \label{fig:rate}
\end{figure}

\subsection{Ablation Study of IKD}
Table~\ref{tab:ab_imp} presents the ablation study on the operations with the Importance-based Knowledge Distillation (IKD).
The experiments are conducted on the ADE20K dataset under the CISS 100-10 scenario.
This study evaluates the impact of three key operations in IKD: \textbf{Importance}, which represents importance of each query on the previous classes, \textbf{Weight}, which determines whether the importance vector is weighted in each step based on the number of classes, and \textbf{Norm}, which denotes whether min-max normalization is applied to the importance vector. 
The 1\textsuperscript{st} row represents the baseline setup, where the distillation importance of all queries are uniformly set to 1.0, without applying either weighting or normalization.
The results reveal the following trends. 
Without any additional operations (1\textsuperscript{st} row), the model achieves mIoU scores of 46.8\%, 26.6\%, and 40.1\% for old classes (1-100), new classes (101-150), and all classes, respectively. 
Applying importance and min-max normalization (3\textsuperscript{rd} row) improves the performance on new classes (27.9\% compared to 26.6\%), resulting in a slight increase in overall mIoU to 40.3\%.
Using importance and weighting (4\textsuperscript{th} row) remarkably enhances the old class mIoU to 48.0\%, while maintaining comparable performance on new classes. 
Finally, combining both weighting and min-max normalization (5\textsuperscript{th} row) achieves the best overall performance, with the mIoU of 41.1\%, including balanced improvements for both old (47.8\%) and new classes (27.7\%).
These results highlight the complementary roles of weighting and normalization in improving the performance of the IKD. 

\begin{table}[!htt]
    \centering
    \resizebox{\linewidth}{!}{
    \begin{tabular}{c|c|c|ccc}
        \toprule
          \multirow{2}{*}{\textbf{Importance}}&\multirow{2}{*}{\textbf{Weight}}& \multirow{2}{*}{\textbf{Norm}}&  \multicolumn{3}{c}{\textbf{100-10} (6 steps)} \\ 
                                &&                       & 1-100 & 101-150 & all \\ \midrule
                                &&                        & 46.8&26.6 &40.1 \\

                                \checkmark&&                        & 47.2&27.5 &40.6 \\ 
                                \checkmark&&\checkmark  & 46.5& 27.9& 40.3\\ 
              \checkmark&\checkmark&              & 48.0& 26.4& 40.8 \\ 
                    \checkmark&\checkmark&    \checkmark& 47.8  & 27.7    & \textbf{41.1}      \\ 
        \bottomrule
    \end{tabular} }
    \caption{Ablation study on operations of the IKD module. The experiments are conducted on the CISS 100-10 task of the ADE20K.
    Note that the ``\checkmark'' denotes whether the operation is utilized.  }
    \label{tab:ab_imp}
\end{table}

\begin{figure*}[!t]
    \centering    \includegraphics[width=0.97\linewidth]{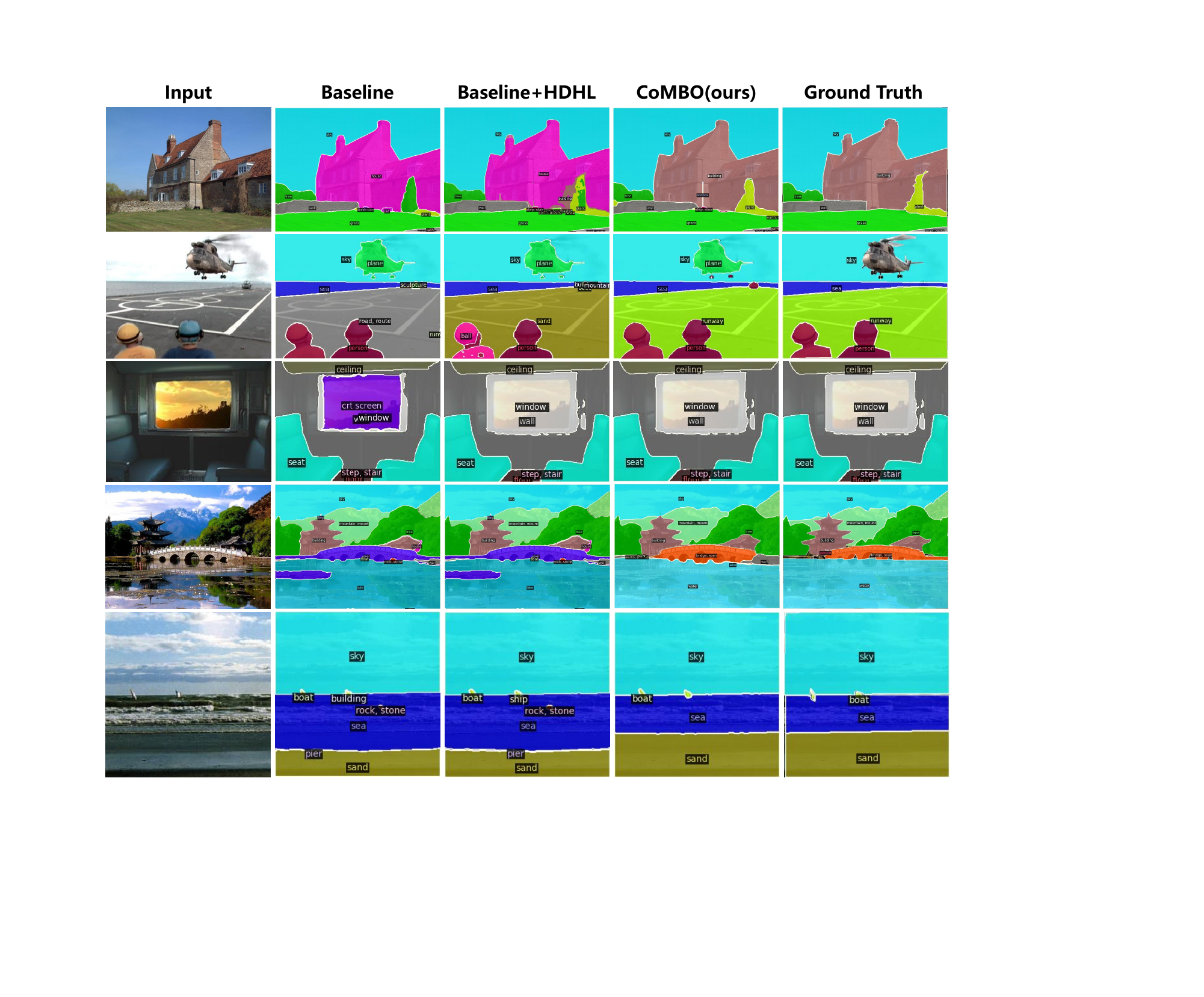}
    \caption{Qualitative results of CoMBO comparing to Baseline, Baseline+HDHL on 100-10  CISS task of ADE20K.
    Each class is uniquely represented by a specific color, making both boundary accuracy and correct color alignment with the ground truth essential for evaluation.}
    \label{fig:sem_qual1}
\end{figure*}
\subsection{Analysis of Class Embeddings}

We introduce the QCR module in Sec.~\ref{sec:qcr}, where the classification prediction from the class embedding $\mathcal{E}_{cls, L-1}$ of layer $L-1$ determines whether using QCR and which QCR should be selected according to the class prediction. 
Therefore, the premise of using the QCR module effectively is that the classification prediction from $\mathcal{E}_{cls, L-1}$ is the same as the classification prediction from $\mathcal{E}_{cls,L}$ of layer $L$. 
Only the inheritable prediction between adjacent layers could enable the class-specific adaptation from the QCR module focusing on its corresponding class. 
We record the proportion of the samples with the same classification prediction results between the $\mathcal{E}_{cls, L-1}$ and $\mathcal{E}_{cls,L}$ w/o QCR module of current classes in Fig.~\ref{fig:rate}. 
The result shows that almost all the results from the corresponding queries meet the requirement on more than 95\% samples, and the average proportion reaches above 98\%. 
These statistics support the premise of our proposed QCR module, and the ablation studies in Sec.~\ref{sec:ab_com} and Sec.~\ref{sec:sp_ab_com} show the effectiveness of QCR module that provides a more harmonious branched optimization structure. 

\subsection{Hyperparameters setting}
We present ablation studies of $\lambda_{KL}$ and $\lambda_{IKD}$ in Tab.~\ref{tab:ab_h}, where we analyze their impact on the CIPS 100-10 task of the ADE20K dataset.
For $\lambda_{KL}$, performance improves as the value increases from 1 to 5, reaching the highest score of 35.6. However, further increasing $\lambda_{KL}$ to 7 and 10 results in a slight decline, suggesting that excessive regularization may restrict model flexibility.
Similarly, for $\lambda_{IKD}$, performance peaks at 35.61 when $\lambda_{IKD}=3$, while larger values (5 and 10) show diminishing returns or slight degradation. 
Additionally, the $\lambda_{cls}$ setting follows Mask2Former~\cite{mask2former}. 
\begin{table}[!ht]
    \centering
    \begin{tabular}{c|cccccc}
    \hline
        $\lambda_{KL}$ & 1 & 3 & 5 & 7 & 10 \\
        \hline
        100-10 & 33.4 & 34.9 & \textbf{35.6} & 35.3  & 34.5 \\
        \hline
        $\lambda_{IKD}$ & 0 & 1 & 3 & 5 & 10 \\
        \hline
        100-10 & 34.88 & 35.38 & \textbf{35.61} & 35.60 & 34.93 \\
        \hline
    \end{tabular}
    \caption{Ablation study on hyperparameter $\lambda_{KL}$ and $\lambda_{IKD}$. }
    \label{tab:ab_h}
\end{table}

\section{Additional Qualitative Results}

\begin{figure*}[!t]
    \centering    \includegraphics[width=\linewidth]{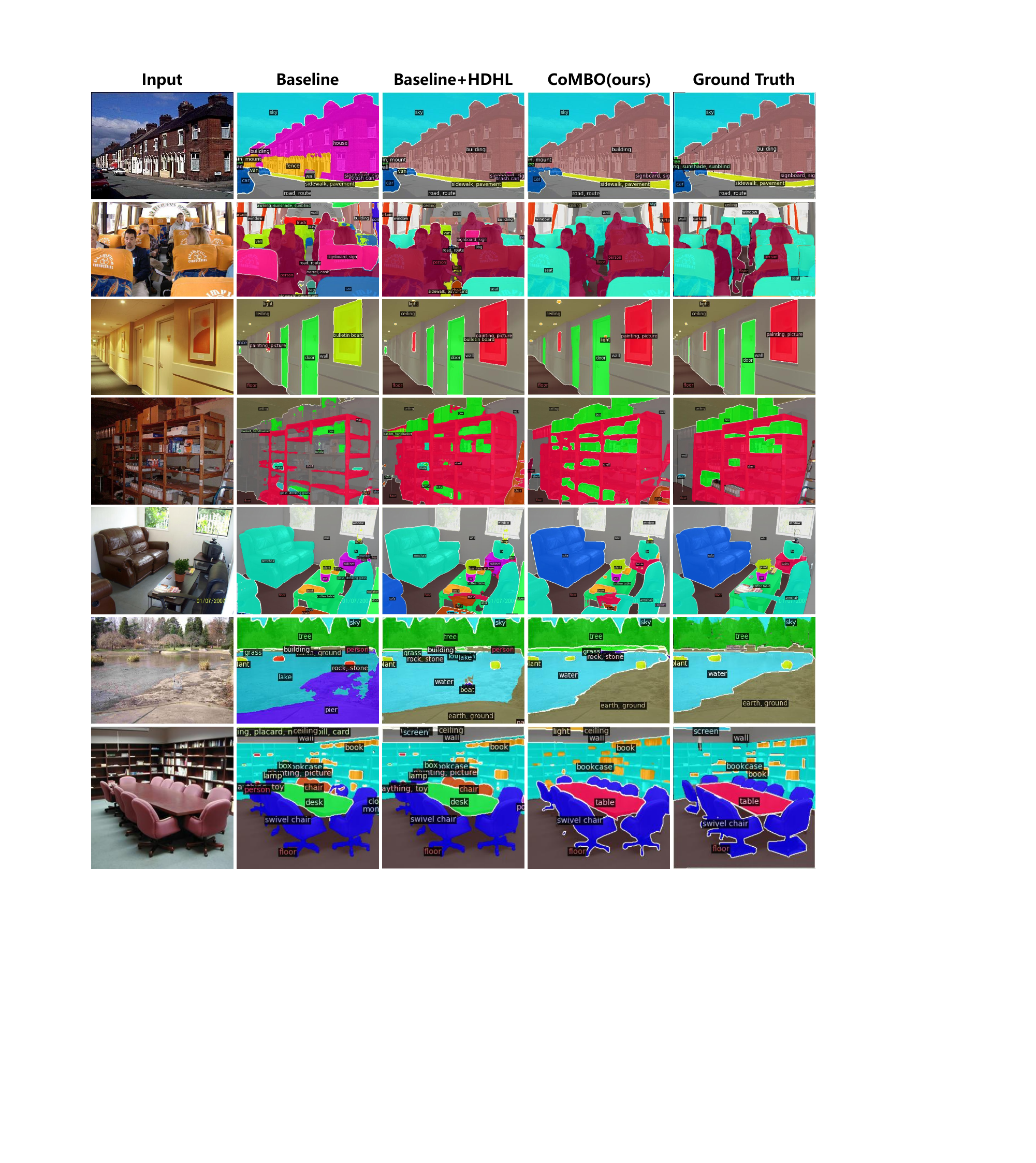}
    \caption{Qualitative results of CoMBO comparing to Baseline, Baseline+HDHL on 100-10  CISS task of ADE20K.
    Each class is uniquely represented by a specific color, making both boundary accuracy and correct color alignment with the ground truth essential for evaluation.}
    \label{fig:sem_qual2}
\end{figure*}

In this section, we perform additional qualitative analysis by contrasting our proposed CoMBO method (3\textsuperscript{rd} column) with both the Baseline (1\textsuperscript{st} column) and Baseline+HDHL (2\textsuperscript{nd} column) on the 100-10 scenario in the CISS, as shown in Fig. \ref{fig:sem_qual1} and Fig.~\ref{fig:sem_qual2}. 
In the 1\textsuperscript{st} and 2\textsuperscript{nd} columns, the Baseline fails to accurately recognize the old classes after the incremental learning, leading to incomplete or incorrect predictions for objects such as \textit{Building} (1\textsuperscript{st} row) and \textit{runway} (2\textsuperscript{nd} row). While Baseline+HDHL shows some improvement in segmenting new classes, it struggles with preserving the masks of initial classes, resulting in the misclassification of \textit{sand} (4\textsuperscript{th} row) and \textit{Bridge} (5\textsuperscript{th} row). 
In contrast, our CoMBO method (3\textsuperscript{rd} column) successfully identifies the incremental classes, such as \textit{stool} (4\textsuperscript{th} row of Fig.~\ref{fig:sem_qual2}), while maintaining accurate predictions for the old classes, as evidenced by the precise segmentation of \textit{Building} (1\textsuperscript{st} row) and \textit{Earth} (7\textsuperscript{th} row of Fig.~\ref{fig:sem_qual2}). 
Additionally, CoMBO achieves finer boundary details for the segments, demonstrating improved refinement capabilities. These results highlight CoMBO's superior performance in reducing the conflict between the retention of old class knowledge and the acquisition of new class information.

\end{document}